\documentclass{article}

\usepackage[preprint]{neurips_2026}

\usepackage[utf8]{inputenc} 
\DeclareUnicodeCharacter{00A0}{~}
\usepackage[T1]{fontenc}    
\usepackage{hyperref}       
\usepackage{url}            
\usepackage{booktabs}       
\usepackage{amsfonts}       
\usepackage{nicefrac}       
\usepackage{microtype}      
\usepackage{xcolor}         
\usepackage{amsmath}
\usepackage{graphicx}
\usepackage{amsthm, amssymb}
\usepackage{cleveref}
\usepackage{subcaption}
\usepackage{tabularx}
\usepackage{enumitem}
\usepackage{caption}
\usepackage{multirow}
\usepackage{float}
\setkeys{Gin}{draft=false}

\title{Subspace Crystallization in Grokking: Rank Compression Tracks the Phase Transition}

\author{%
  Shlok Mehendale\thanks{Corresponding author: \texttt{shlok.mehendale@lossfunk.com}}\\
  Lossfunk\\
  \And
  Paras Chopra\\
  Lossfunk\\
}

\begin{document}

\thispagestyle{empty}
\begin{center}
\vspace*{-1cm}
\includegraphics[width=0.25\textwidth]{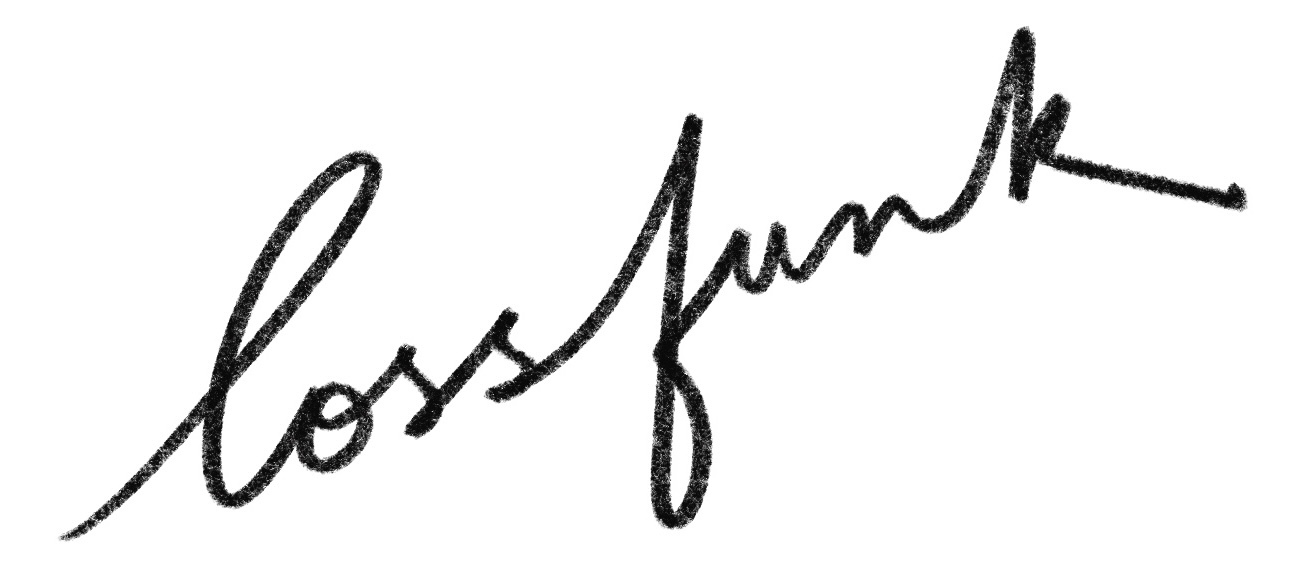}
\vspace{0.3cm}
\end{center}

\maketitle

\begin{abstract}
Grokking describes understanding something deeply and intuitively. In neural networks, it is the phenomenon where a network generalizes long after memorizing its training data. This delay has been linked to a fall in the effective rank of the network's weight matrices, but whether this rank compression drives grokking or follows it is still unresolved. We study a mean-field MLP and a one-layer transformer trained on modular arithmetic and observe that:
(i)~\emph{Compression follows the onset, not memorization.} Every transformer memorizes its training data within 250 steps, while the grokking onset (the first step at which test accuracy reaches 95\%) varies from 6,500 to 21,000 steps. The steepest fall in effective rank comes about 900 steps after the onset, however late the onset occurs, and networks that memorize without grokking compress far less.
(ii)~\emph{Subspace crystallization.} We call the span of the fewest top singular directions of a weight matrix that preserve 99\% of test accuracy its \emph{sufficient subspace}. Its size can shrink as the margin between the correct and the best wrong output grows, so we also count the directions needed to keep the outputs within 10\%. In the transformer, this count falls during the transition from a quarter or more of the available directions to about a tenth, which we call \emph{subspace crystallization}.
(iii)~\emph{Imposing low rank does not speed up grokking.} Against a factorized full-rank control, capping the rank of the transformer's MLP input weights delays the onset, and capping the rank of its embedding stops 15 of 20 runs from grokking.
Compression is therefore a \textit{signature} of grokking rather than its cause.
\end{abstract}

\section{Introduction}
Grokking~\citep{power2022grokking}, memorization followed by delayed generalization, offers a rare window into how circuits emerge from gradient descent: a clean phase transition from a lazy, memorizing regime to a rich, generalizing one~\citep{kumar2024grokking}. Substantial progress exists on \emph{what} grokking networks compute \citep{nanda2023progress,chughtai2023toy,gromov2024grokking}, and a recurring observation is that the effective rank (ER) of their weight matrices falls as test accuracy jumps \citep{liu2023complexity,demoss2024complexity,lei2026geometric}. This has led to proposals that a pressure toward low-rank, small-norm or efficient circuits causes grokking \citep{nanda2023progress,liu2023omnigrok,varma2023circuit}. We ask two questions. \textbf{(Q1)} Does this rank compression cause grokking, or follow it? \textbf{(Q2)} Does the learned solution lie in the subspace spanned by a few leading directions of each weight matrix?

\begin{figure}[!t]
  \centering
  \begin{subfigure}[b]{\linewidth}
    \centering
    \includegraphics[width=0.9\linewidth]{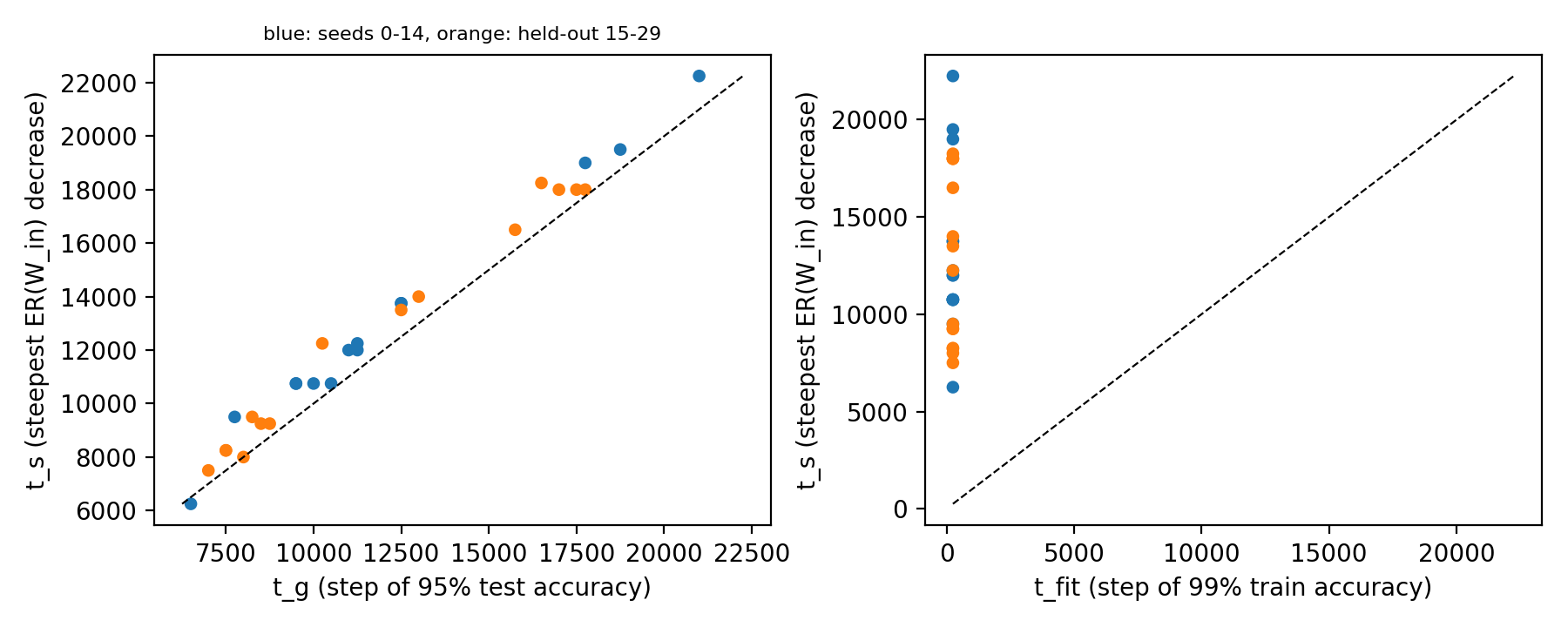}
    \caption{Step of the steepest effective-rank decrease of $W_\text{in}$ against the grokking onset (left) and the step at which training accuracy reaches 99\% (right). Blue: seeds 0--14. Orange: held-out seeds 15--29. Dashed line: identity.}
    \label{fig:ts_tg}
  \end{subfigure}
  \begin{subfigure}[b]{\linewidth}
    \centering
    \includegraphics[width=0.9\linewidth]{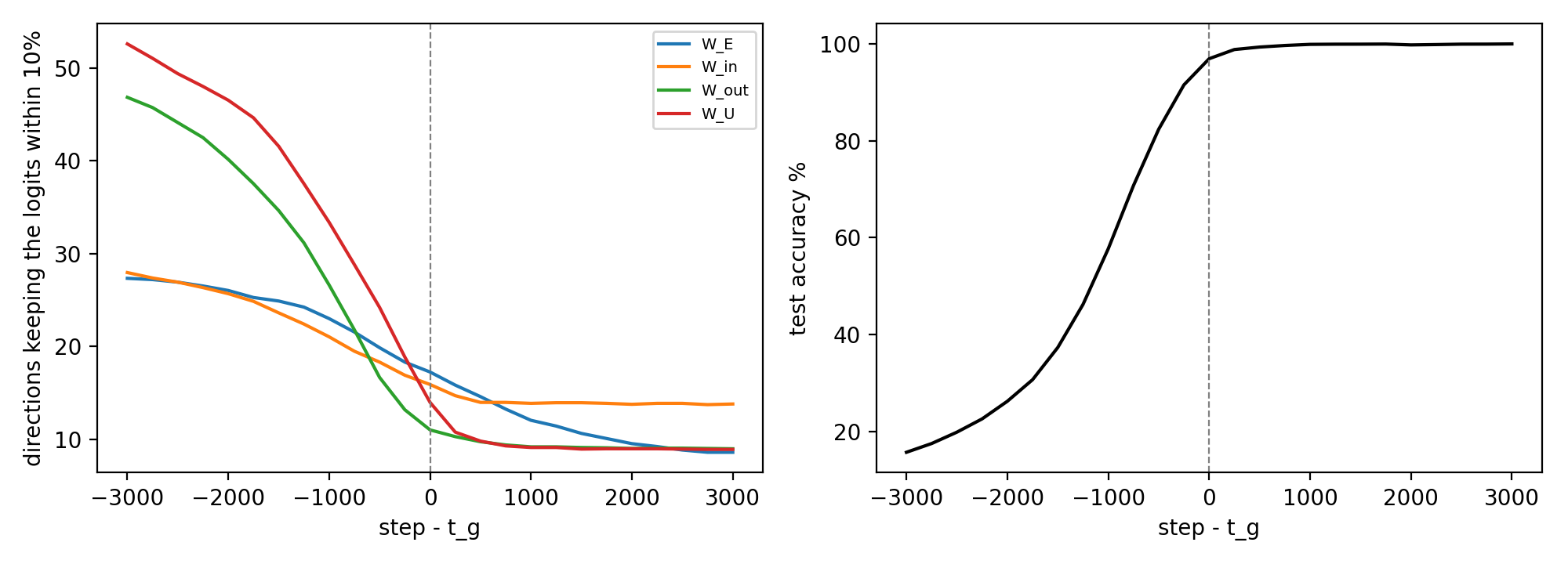}
    \caption{Number of directions of each weight matrix needed to keep the network's outputs within 10\% (left) and test accuracy (right), with each seed aligned at its own onset $t_g$.}
    \label{fig:tf_dense}
  \end{subfigure}
  \caption{\textbf{Compression follows the grokking onset, and the directions needed to keep the outputs fall during the transition} (one-layer transformer, 29 grokked seeds). (a)~Every seed fits its training data by step 250, yet compression follows the grokking onset across a threefold range of onset times. (b)~The number of directions needed to keep the outputs within 10\% falls during the rise in test accuracy and settles at 8--13 per matrix.}
  \label{fig:teaser}
\end{figure}

The usual evidence cannot settle these questions, for three reasons. First, rank compression is usually observed after two events, memorization (the first step at which training accuracy reaches 99\%) and the rise in test accuracy (Figure~\ref{fig:teaser}a), so the timing alone cannot explain which of the two it follows. Second, timing without interventions cannot show cause. Lastly, the compactness of a solution is naturally measured by truncating each weight matrix to lower and lower rank until test accuracy drops. Keeping a weight matrix’s $k$ strongest singular directions gives the closest rank-$k$ matrix, which acts only within their span. The span of the fewest top directions that leave the test accuracy unchanged, which we call the \emph{sufficient subspace}, is then taken as the subspace the solution needs, analogous to the intrinsic dimension of \citet{li2018intrinsic}. Accuracy, however, only checks which output is largest. The margin between the correct-class output and the largest other output sets how far the outputs can move before a prediction changes. As margins grow, more directions can be deleted before any prediction changes, so the count can fall even though the network’s outputs still depend on those directions.

We address these questions with geometric metrics (effective rank, stable rank, sparsity, inverse participation ratio) and multiple interventions (truncation of weight matrices to their top, bottom or random singular directions, complement interventions, transplant patching with orthogonal alignment, a nuclear-norm penalty, rank caps compared with factorized controls, and optimizer swaps). We study modular addition $(a{+}b)\bmod 97$ in two architectures with multiple random initializations. In the transformer, we analyze the embedding $W_E$, the unembedding $W_U$, and the input and output matrices $W_\text{in}$ and $W_\text{out}$ of its MLP block. We call the first step at which test accuracy reaches 95\% the \textit{grokking onset}. To separate memorization from the onset, we follow the transformer with a checkpoint every 250 steps and compare networks that grok with networks that memorize but never grok. To count directions without the margin, we judge each truncation by how much it changes the network's outputs. To test cause, we limit the rank of a weight matrix throughout training and compare each limit with a factorized full-rank control. We test the timing results on held-out transformer seeds and on two further tasks, $ab\bmod 97$ and $(a{+}b)\bmod 113$. Our results point to compression as a \textit{signature} of grokking rather than its cause.

\paragraph{Contributions}
\begin{itemize}[leftmargin=*,itemsep=2pt,topsep=2pt]

    \item \textbf{Compression follows grokking, not memorization} (\S\ref{sec:causal}). Every transformer memorizes its training data within 250 steps, but the steepest fall in the effective rank of $W_\text{in}$ comes about 900 steps after the grokking onset, whether the onset occurs at step 6,500 or 21,000. Networks that memorize without grokking lower this effective rank by 15--23\%, against $60\pm5\%$ in those that grok. In the MLP, rank compression follows the onset in all 45 runs across three tasks. The timing replicates on 15 held-out seeds and, in the MLP, on two further tasks.

    \item \textbf{Subspace crystallization: the transformer’s solution narrows to a low-dimensional subspace} (\S\ref{sec:crystallisation}, \S\ref{sec:transformer}). Before the transition, no weight matrix can be truncated below 27--53 directions without changing the network’s outputs by more than 10\%. After the transition, 8--13 directions suffice (Figure~\ref{fig:teaser}b). We call this narrowing of the subspace that the outputs depend on \emph{subspace crystallization}. It is consistent with the removal of memorizing components described by \citet{nanda2023progress}. We judge truncations by the outputs rather than only by the test accuracy. In the MLP trained with mean-squared error, truncation to 65 of 128 directions keeps the test accuracy, but keeping the outputs within 10\% needs 122--127.

    \item \textbf{Imposing low rank before the transition does not speed up grokking} (\S\ref{sec:bottleneck}). If rank compression drove grokking, removing directions in advance should help. Instead, against a factorized full-rank control, capping $W_\text{in}$ at rank 32, 24 or 16, all above the 13 directions its solution needs, never lowers the mean onset, and capping $W_E$ prevents grokking in 15 of 20 runs. Compressing the MLP's first layer with a nuclear-norm penalty before the onset also delays grokking, and comparing with the unfactorized network instead of the factorized control would wrongly suggest that low rank speeds grokking up, because the factorization changes the onset by itself.

\end{itemize}

We also report three secondary results. The late fall of effective rank in the MLP is set by AdamW's $\epsilon$: once the square root of its second-moment estimate falls below $\epsilon$, its updates reduce to gradient descent with momentum and learning rate $\text{lr}/\epsilon$ (\S\ref{sec:fourstage}). Different MLP seeds share their hidden representation up to an orthogonal rotation, but not with networks trained on another task or on random labels (\S\ref{sec:universality}). The Fourier inverse participation ratio (IPR), which measures how periodic the hidden units are \citep{doshi2024togrok}, rises before the onset but also in every network that never groks, so it cannot serve as a label-free early-warning signal (\S\ref{sec:ipr}). Table~\ref{tab:claims_scope} summarizes the evidence for the three contributions in both architectures.


\section{Related Work}
\label{sec:related}

\paragraph{Grokking.}
Grokking was given its first mechanistic account by \citet{nanda2023progress}, who identified the learned algorithm as a sparse Fourier cosine decomposition, formed in phases of memorization, circuit formation and cleanup. \citet{chughtai2023toy} extended this to general group composition, and \citet{gromov2024grokking} established exact solvability with analytic weights.
Proposed causes of the delay include circuit efficiency \citep{varma2023circuit}, the weight norm \citep{liu2023omnigrok}, softmax collapse \citep{prieto2025grokking} and a transition from lazy to rich training \citep{kumar2024grokking}. 
\citet{lei2026geometric} describe grokking as construction followed by compression, tracked through the network's Jacobian, consistent with our timing results. We add the separation of compression from memorization, negative controls and a replication on held-out seeds and tasks. \citet{musat2025geometry} proves that, in the limit of small learning rate and weight decay, gradient descent after memorization minimizes the weight norm on the zero-loss manifold. Our MLP lies outside this regime: late in training, AdamW acts as gradient descent with a large learning rate $\text{lr}/\epsilon = 10^5$, and the weight norm grows (\S\ref{sec:fourstage}).

\paragraph{Complexity measures of grokking.}
Network complexity falls around the accuracy jump, whether measured by the linear mapping number \citep{liu2023complexity} or by the compressed description length of the weights \citep{demoss2024complexity}. We test whether this fall drives grokking or follows it, and show that a count of directions judged by accuracy can register compression that the network’s outputs do not show (\S\ref{sec:crystallisation}).

\begin{table}[tbp]
\centering
\footnotesize
\setlength{\tabcolsep}{5pt}
\renewcommand{\arraystretch}{1.15}
\caption{\textbf{Evidence for the three contributions in both architectures.} Effective rank is measured on $W_1$ (MLP) and $W_\text{in}$ (transformer). Output counts are the fewest directions per matrix that change the network's outputs by at most 10\%.}
\label{tab:claims_scope}
\begin{tabularx}{\linewidth}{>{\raggedright\arraybackslash}p{4.4cm}>{\raggedright\arraybackslash}X>{\raggedright\arraybackslash}X}
\toprule
 & MLP & Transformer \\
\midrule
\multicolumn{3}{l}{\textit{1. Compression follows grokking, not memorization}} \\
Compression starts after test accuracy rises & 45 of 45 runs & 28 of 29 seeds \\
Fall in effective rank & 10\% if grokked & $60\pm5\%$ if grokked, 15--23\% if not \\
\midrule
\multicolumn{3}{l}{\textit{2. Subspace crystallization: the transformer’s solution contracts into a low-dimensional subspace as it groks}} \\
Directions needed per matrix & Accuracy: from 110 to 65 of 128\newline Outputs: 122--127 of 128 (MSE) & Outputs: from 27--53 to 8--13, during the transition \\
\midrule
\multicolumn{3}{l}{\textit{3. Imposing low rank does not speed up grokking}} \\
Rank caps against a factorized control (Table~\ref{tab:tf_caps}) & Caps delay the onset & $W_\text{in}$: onset unchanged or delayed\newline $W_E$: 15 of 20 runs fail to grok \\
\bottomrule
\end{tabularx}
\end{table}

\paragraph{Low-rank bias and factorized controls.}
Gradient descent is biased toward low-rank solutions \citep{huh2023lowrank} and moves within a small subspace of parameter space \citep{gurariari2018gradient}. Factorizing a matrix strengthens this bias \citep{arora2019implicit} and accelerates optimization \citep{arora2018acceleration}, so a rank cap written as a product $W=AB$ changes the optimization as well as the rank. We therefore compare each cap with a factorized full-rank control (\S\ref{sec:bottleneck}).

\paragraph{Sparse circuits and universality.}
\citet{merrill2023tale} describe grokking as a competition between a sparse generalizing subnetwork and a dense memorizing one, consistent with the few directions per matrix that the transformer’s solution keeps. Networks trained on modular arithmetic can learn different algorithms \citep{zhong2023clock} or share one across architectures \citep{chughtai2023toy}. Our MLP seeds use different Fourier coordinates but share their hidden representation up to a rotation (\S\ref{sec:universality}). Appendix~\ref{app:related_extended} discusses further work.

\section{Setup}
\label{sec:setup}

\paragraph{Architecture 1: mean-field MLP.}
A single-hidden-layer MLP with 128 hidden units and quadratic activations, $f(x)=\tfrac{1}{128}W_2\,\phi\big(W_1x/\sqrt{194}\big)$ with $\phi(z)=z^2$ and input $x=[e_a;e_b]\in\mathbb{R}^{194}$, the concatenation of two one-hot vectors ($W_1\in\mathbb{R}^{128\times194}$, $W_2\in\mathbb{R}^{97\times128}$, i.i.d.\ $\mathcal{N}(0,1)$ initialization). The quadratic activation makes the Fourier computation explicit: for $x=\cos\theta$ it doubles frequencies, $x^2=\tfrac12(1+\cos2\theta)$, and applied to a sum of operand features it produces the products $\cos\theta_a\cos\theta_b$ from which $\cos(\theta_a+\theta_b)$ is read out. We train it with mean-squared error to one-hot targets, as in \citet{gromov2024grokking}, using full-batch AdamW (lr $10^{-3}$, weight decay $10^{-2}$, $\epsilon=10^{-8}$) on a 35\% train split for $10^5$ steps, with 15 seeds. Five MLPs trained on randomly permuted labels serve as negative controls, and for the replication the MLP is also trained on $ab\bmod 97$ and $(a{+}b)\bmod 113$ (15 seeds each).

\paragraph{Architecture 2: one-layer transformer.}
Four attention heads, $d_\text{model}=128$, $d_\text{head}=32$, $d_\text{mlp}=512$, ReLU MLP, no layer normalization and biases fixed at zero, following \citet{nanda2023progress}. The inputs are $[a,b,{=}]$ and the prediction is read at the final position. We train it with cross-entropy, using full-batch AdamW (lr $10^{-3}$, weight decay $1.0$, $\beta=(0.9,0.98)$) on a 30\% train split for 25,000 steps, with 15 seeds. Fourteen of these seeds reach 95\% test accuracy within this budget. The remaining seed, and three seeds each trained with weight decay 0.1 and 0.3, memorize the training set without generalizing and serve as negative controls. Seeds 15--29, trained only after the timing measure had been defined on seeds 0--14, form the held-out set of the replication, and all of them grok. For the time-resolved analyses, every run was repeated with a checkpoint every 250 steps. We analyze the embedding $W_E$, the unembedding $W_U$, the MLP projections $W_\text{in}$ and $W_\text{out}$, and the attention projections $W_Q$, $W_K$, $W_V$, $W_O$ (heads stacked).

\paragraph{Timing.}
The \textbf{grokking onset} $t_g$ is the first logged step with test accuracy $\geq95\%$, and the \textbf{memorization step} $t_\text{fit}$ the first logged step with training accuracy $\geq99\%$. A run counts as grokked if its final test accuracy is at least 95\%. Runs described as never grokking did not reach this threshold within their training budget ($10^5$ steps for the MLP, 25,000 for the main transformer runs). Throughout the paper, ``before'' and ``after generalization'' mean before and after $t_g$.

\paragraph{Effective rank.}
Our primary metric is the \textbf{effective rank} $\text{ER}(W)=\exp(H(\hat{\boldsymbol{\sigma}}))$ \citep{roy2007effective}, the exponential of the entropy of the normalized singular values $\hat\sigma_i=\sigma_i/\sum_j\sigma_j$. A matrix whose $k$ nonzero singular values are equal has $\text{ER}=k$. Appendix~\ref{app:setup} defines the other tracked metrics (stable rank, sparsity, spectral IPR, weight and gradient norms), and Section~\ref{sec:ipr} the Fourier IPR.

Code to reproduce all experiments is available at \url{https://github.com/Lossfunk/Causal-geometric-grokking}.

\section{The Four-Stage Training Trajectory (MLP)}
\label{sec:fourstage}

\begin{figure}[!t]
  \centering
  \includegraphics[width=0.95\linewidth]{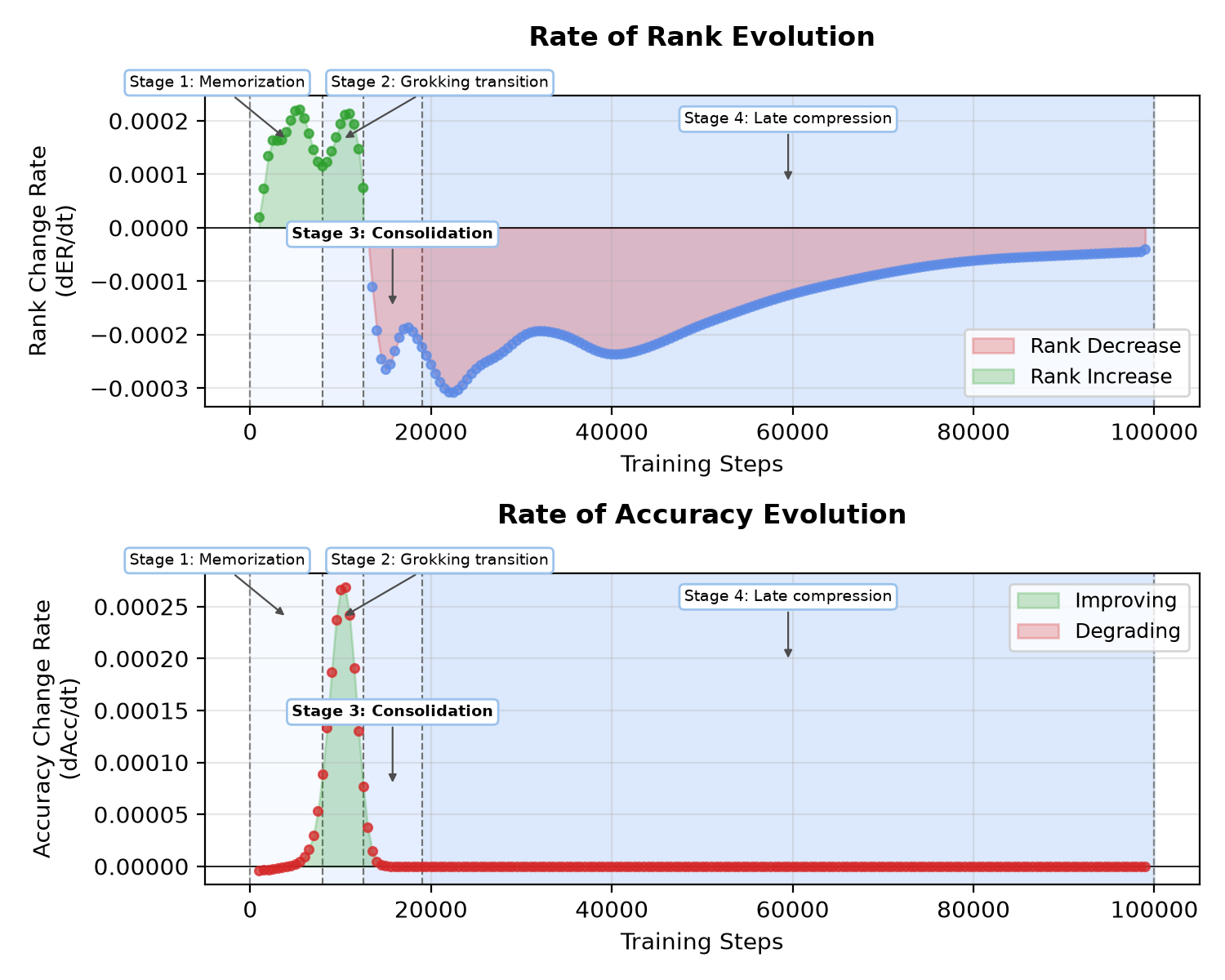}
    \caption{\textbf{The four training stages of the MLP} (seed 0): rate of change of the effective rank of $W_1$ (top) and of test accuracy (bottom), computed from the five-step moving average used for the steepest-decrease measure (Appendix~\ref{app:tf_timing}). Stage~2 starts at the first step with test accuracy above 5\%, Stage~3 at $t_g$ and Stage~4 at $t_c$. Effective rank grows during memorization and the grokking transition, turns negative at $t_g$ and keeps falling through Stages~3 and~4.}
  \label{fig:stages}
\end{figure}

Figure~\ref{fig:stages} shows the four stages through the rates of change of effective rank and test accuracy. Numbers are means $\pm$ s.d.\ over 15 seeds, and Appendix Figure~\ref{fig:training_dynamics} shows all training metrics.

\paragraph{Stage~1 (Memorization):} training accuracy rises to $97.8\pm0.4\%$ by step 8,000 while test accuracy stays below 5\%. Test loss rises before it falls, a double-descent signature \citep{davies2023unifying}. Effective rank rises slightly, from $114.6$ for the random initial $W_1$ to about $116$.
\paragraph{Stage~2 (Grokking transition):} test accuracy rises from $37.0\pm8.0\%$ at step 10,000 to 95\% at $t_g=12{,}633\pm297$, and training accuracy reaches 100\%. Effective rank rises only slightly ($115.7\pm0.4$ at step 10,000, $115.9\pm0.6$ at step 12,500), while the hidden units become periodic in their inputs, as measured by the Fourier IPR (\S\ref{sec:ipr}), and the gradient norm peaks just after $t_g$.
\paragraph{Stage~3 (Consolidation):} test accuracy is 100\% and ER stays near its maximum until the compression onset $t_c=19{,}000\pm3{,}279$, the first step at which it falls below 99\% of its running maximum. Over the same interval the median margin grows from $0.105\pm0.017$ at step 12,500 to $0.470\pm0.019$ at step 20,000, and the directions of $W_1$ needed to keep 99\% of test accuracy fall from $109.5\pm3.2$ to $64.9\pm1.7$, a consequence of the growing margin (\S\ref{sec:crystallisation}). The stable rank of $W_1$ peaks around $t_g$ and starts to fall.
\paragraph{Stage~4 (Late compression):} ER falls to $104.0\pm2.1$ by step $10^5$, and $99\pm2\%$ of its total decrease happens after $t_g$. The stable rank keeps falling, the weight norms grow throughout and the margin barely changes ($0.599$ to $0.611$ between steps 50{,}000 and $10^5$). Observing this stage requires training well past the point where test accuracy saturates.

\paragraph{Stage~4 is set by the effective learning rate.}
Stage~4 is growth of the leading singular directions of $W_1$ against weight decay: between steps 20{,}000 and $10^5$ the largest singular value doubles (112 to 218) while the remaining 63 directions keep their size (Appendix~\ref{app:causal_dim}). The loss gradient that drives this growth is tiny ($3.4\times10^{-6}$ at step 20,000), points increasingly outward along the weights (its cosine with $-\theta$ rises from 0.19 to 0.89) and concentrates in the top 65 directions of $W_1$ (from 43\% to 87\% of its squared norm). For 87--100\% of the coordinates, the square root of Adam's second-moment estimate lies below $\epsilon$, so the update $m/(\sqrt{v}+\epsilon)$ is close to $m/\epsilon$: AdamW acts as gradient descent with learning rate $\text{lr}/\epsilon=10^5$ (Appendix~\ref{app:adam}). Continuing the main runs from step 20{,}000 with other optimizers confirms that this effective learning rate, not Adam itself, governs Stage~4 (Table~\ref{tab:optimizers}). At effective learning rates of $10^3$ and $10^4$, AdamW and gradient descent reach the same weights. At $10^5$, gradient descent reproduces most of Stage~4 and SGD with momentum reproduces all of it (effective rank $108.3$ and $103.8$, against $104.8$ under AdamW). At $10^3$, weight decay shrinks the norm and effective rank stays at 117.8, while $\epsilon\leq10^{-9}$ strengthens Stage~4.

\begin{table}[tbp]
\centering
\small
\setlength{\tabcolsep}{4pt}
\caption{\textbf{Stage~4 tracks the effective learning rate.} Main seeds 0--2 are continued from step 20,000 to $10^5$ with each optimizer, and entries are means. The effective learning rate is lr for gradient descent, lr$/(1-0.9)$ for SGD with momentum 0.9, and lr$/\epsilon$ for AdamW ($\text{lr}=10^{-3}$). For $\epsilon=10^{-9}$, lr$/\epsilon$ is an upper bound, since $\sqrt{\hat v}$ is comparable to $\epsilon$. Weight decay is set so that the per-step shrinkage matches AdamW’s. $d_{0.99}$: fewest directions that keep 99\% of test accuracy.}
\label{tab:optimizers}
\begin{tabular}{lccccc}
\toprule
Optimizer & Effective lr & $\|W_1\|_F$ & $\sigma_1(W_1)$ & ER$(W_1)$ & $d_{0.99}$ \\
\midrule
Start (step 20,000) & -- & 726 & 111 & 115.1 & 65 \\
\midrule
AdamW, $\epsilon=10^{-6}$ & $10^{3}$ & 658 & 85 & 117.8 & 65 \\
Gradient descent & $10^{3}$ & 662 & 85 & 117.8 & 65 \\
\addlinespace
AdamW, $\epsilon=10^{-7}$ & $10^{4}$ & 775 & 120 & 114.8 & 70 \\
Gradient descent & $10^{4}$ & 783 & 126 & 114.6 & 72 \\
\addlinespace
AdamW, $\epsilon=10^{-8}$ (main) & $10^{5}$ & 1{,}040 & 203 & 104.8 & 89 \\
Gradient descent & $10^{5}$ & 833 & 148 & 108.3 & 87 \\
SGD with momentum 0.9 & $10^{5}$ & 1{,}093 & 216 & 103.8 & 91 \\
\addlinespace
AdamW, $\epsilon=10^{-9}$ & $10^{6}$ & 1{,}582 & 318 & 94.7 & 98 \\
\midrule
AdamW, $\epsilon=10^{-8}$, no weight decay & $10^{5}$ & 1{,}257 & 261 & 97.8 & 94 \\
\bottomrule
\end{tabular}
\\[2pt]
{\footnotesize Full sweep in Appendix~\ref{app:adam}. Gradient descent at $10^6$ diverges.}
\end{table}

\paragraph{Weight decay, loss and activation.}
The MLP (with MSE loss) groks without weight decay: from weight decay 0 to 0.1, $t_g$ barely moves ($12{,}700\pm447$ to $13{,}400\pm418$), but the Stage-4 decrease in effective rank shrinks from $14.9\pm1.9\%$ to $0.5\pm0.4\%$ ($9.7\pm1.5\%$ at the default $10^{-2}$), because weight decay restrains the growth of the leading directions. Trained with cross-entropy at learning rates $3\times10^{-3}$ and $10^{-2}$, the MLP groks in every seed ($t_g=27{,}500\pm2{,}372$ and $10{,}100\pm962$) and shows Stage-4 decreases of $10.9\pm0.5\%$ and $37.9\pm1.9\%$. At lr $10^{-3}$ it groks in only one of five seeds within $10^5$ steps. ReLU and GELU MLPs also grok, later than quadratic ones, with periodic first-layer weights and Stage-4 decreases of 14--17\% (Appendix~\ref{app:variants}). With the same hyperparameters, the quadratic MLP groks 2.6 to 3.5 times sooner than ReLU and GELU MLPs, so the quadratic activation speeds up grokking in our setting but is not required for the MLP to grok (Appendix~\ref{app:nonlinearity}).


\section{Subspace Crystallization: Locating the Sufficient Subspace}
\label{sec:crystallisation}

This section defines the sufficient subspace and the output count $d_\text{out}$, and uses them to test whether the MLP crystallizes. The transformer's crystallization is shown in Section~\ref{sec:transformer}.

\paragraph{Definitions.}
We replace a weight matrix $W=U\Sigma V^\top$ by its top-$k$ truncation $W_{[k]}=U_{:k}\Sigma_{:k}V_{:k}^\top$ and leave the rest of the network unchanged. $d_{0.99}(W)$ is the smallest $k$ that keeps 99\% of the test accuracy, and the span $\mathcal{C}$ of the top-$d_{0.99}$ left singular vectors is the \emph{sufficient subspace}. Because accuracy only checks which output is largest, $d_{0.99}$ can fall as margins grow. $d_\text{out}(W)$ is the smallest $k$ with $\|f_{[k]}-f\|_F\leq0.1\,\|f\|_F$ over the test inputs, where $f$ stacks the network's outputs and, for cross-entropy models, each input's mean logit is removed, since softmax ignores it. \emph{Subspace crystallization} is a narrowing of $d_\text{out}$. The leading directions carry the solution: truncation to the top $d_{0.99}$ directions keeps at least 99.26\% of test accuracy, while random subspaces of equal rank keep 22.7\% in the MLP and at most 12.5\% in the transformer (Table~\ref{tab:controls}).

\paragraph{In the MLP, crystallization is only apparent.}
With mean-squared error, crystallization is only apparent. The sufficient subspace of $W_1$ narrows right after the onset, from $d_{0.99}=109.5\pm3.2$ at step 12,500 to $64.9\pm1.7$ at step 20,000, and widens again to $89.1\pm2.3$ by step $10^5$ (Table~\ref{tab:dc_time}). Over the narrowing, however, the median margin grows more than fourfold and $d_\text{out}$ does not fall: it is $122.4\pm0.5$ before the onset, $122.3\pm0.9$ at step 20,000 and $127.3\pm0.6$ at the end (Table~\ref{tab:margin_free}), above the $116.4\pm0.5$ of an untrained network and the $123.0$ of a network trained on random labels. The same holds on both replication tasks (Appendix~\ref{app:baselines}). The narrowing of $d_{0.99}$ therefore measured how far the predictions were from flipping, not how many directions the function uses. Section~\ref{sec:comparison} attributes the high $d_\text{out}$ to the loss: mean-squared error to one-hot targets requires reproducing every Fourier component of the outputs. The later widening of the sufficient subspace comes from the growth of the leading singular directions, which makes the trailing directions matter more for the outputs (Appendix~\ref{app:causal_dim}). The complement $\mathcal{C}^\perp$ is unused for the predictions but is not inert: Gaussian noise added inside $\mathcal{C}^\perp$ lowers accuracy more than the same noise inside $\mathcal{C}$ (71\% against 92\%, Appendix~\ref{app:interventions}). With cross-entropy, which only requires the correct output to be largest, the MLP does crystallize, but late: $d_\text{out}\approx120$ just before and just after the onset, and $81.0\pm1.6$ (learning rate $3\times10^{-3}$) and $69.6\pm2.2$ ($10^{-2}$) at the end of training (Table~\ref{tab:margin_free}).

\begin{table}[tbp]
\centering
\small
\setlength{\tabcolsep}{4pt}
\caption{\textbf{$d_\text{out}$ and $d_{0.99}$ in the MLP} (directions of $W_1$ out of 128, on test inputs unless noted, mean $\pm$ s.d.). $d_\text{out}$: fewest directions that keep the outputs within 10\%. $d_{0.99}$: fewest directions that keep 99\% of test accuracy. Cross-entropy (CE) models use centered logits.}
\label{tab:margin_free}
\begin{tabular}{llccc}
\toprule
Model & Checkpoint & Outputs within 10\% & 99\% of accuracy & Median margin \\
\midrule
MSE (15 seeds) & step 0 (untrained) & $116.4\pm0.5$ & -- & $-0.004$ \\
 & step 11{,}000 (before onset) & $122.4\pm0.5$ & $119.3\pm3.2$ & $0.025$ \\
 & step 20{,}000 & $122.3\pm0.9$ & $64.9\pm1.7$ & $0.470$ \\
 & end ($10^5$) & $127.3\pm0.6$ & $89.1\pm2.3$ & $0.611$ \\
MSE, random labels (5) & end, training inputs & $123.0\pm0.0$ & -- & $0.110$ \\
\midrule
CE, lr $3\times10^{-3}$ (5) & last before onset & $121.4\pm0.9$ & $112.8\pm5.5$ & $3.8$ \\
 & first after onset & $119.0\pm2.2$ & $91.8\pm15.7$ & $5.9$ \\
 & end & $81.0\pm1.6$ & $50.6\pm2.3$ & $9.4$ \\
CE, lr $10^{-2}$ (5) & last before onset & $120.2\pm0.8$ & $104.6\pm4.1$ & $4.2$ \\
 & first after onset & $119.8\pm0.4$ & $98.6\pm4.2$ & $4.9$ \\
 & end & $69.6\pm2.2$ & $48.4\pm2.5$ & $10.2$ \\
\bottomrule
\end{tabular}
\end{table}

\paragraph{Equivalent solutions across random initializations.}
\label{sec:universality}
MLP seeds implement the same solution in different coordinates. Each seed spreads its first-layer Fourier power over most of the 48 frequencies ($45.2\pm0.6$ hold 99\%), with different strongest frequencies (Jaccard overlap $0.05$ over 105 seed pairs), and the top-60 weight subspaces of different seeds overlap no more than random subspaces ($0.466\pm0.027$ against $0.487\pm0.005$ in mean cosine of the principal angles, Appendix~\ref{app:overlap}). Yet after an orthogonal Procrustes rotation fitted on training inputs \citep{schonemann1966generalized}, one seed's readout decodes another seed's post-activations with $95.8\pm2.4\%$ accuracy, against $1.0\%$ without the rotation (Table~\ref{tab:cka}). The rotation does not rescue a random network, a network trained on random labels or a network grokked on $ab\bmod97$ (0.6--1.4\%). Each seed thus implements the same task-relevant representation in a different orientation, consistent with \citet{Schiffman}, who find invariant algorithmic cores in independently trained transformers.

\section{Timing and Cause of Compression}
\label{sec:causal}

We test whether compression drives grokking with three kinds of evidence: its timing across seeds, how much the effective rank decreases in networks that memorize without grokking, and interventions that impose low rank from initialization.

\subsection{Compression Follows the Onset, Not Memorization}
\label{sec:fit}

The transformer separates memorization from the transition, and compression follows the transition. Every one of the 29 grokked seeds (seeds 0--14 and the held-out seeds 15--29) fits its training set within the first 250 steps, while their onsets range from 6{,}500 to 21{,}000 steps. With weight decay 1.0, AdamW multiplies the weights by 0.999 per step, so by step 4,000 the initial weights have decayed to 2\% of their size. Because random weights spread over all directions, this decay alone lowers the effective rank of $W_\text{in}$, to a plateau at about 60\% of its initial value, independently of grokking (Figure~\ref{fig:tf_aligned}). We therefore measure compression only after step 4,000. After this decay, a second, sharp drop follows. The step $t_s$ at which this second drop is steepest follows the onset almost exactly (Pearson correlation $r=0.99$, slope $1.02$, 95\% confidence interval $0.98$--$1.07$), scattering by 506 steps around $t_g$ against 4{,}234 around the memorization step (Figure~\ref{fig:ts_tg}). If compression followed memorization, $t_s$ would cluster at a fixed step.

Networks that memorize without grokking compress far less. From step 4{,}000 to the end of training, the ER of $W_\text{in}$ falls by $60\pm5\%$ in grokked transformers but by only 15--23\% in the seven that never grok, at a peak rate two to five times lower (Appendix~\ref{app:fit}). In the MLP, whose onset and memorization time barely vary across seeds, the random-label networks reach 90\% training accuracy without grokking, and their effective rank falls by $1.1\pm0.3\%$, against $10.5\pm1.8\%$ in grokked MLPs. Comparing seeds at the same training step shows that low rank is acquired at the transition itself (Appendix Table~\ref{tab:matched}): before their own onset, grokked seeds have as many directions as the seed that never groks, and after it far fewer.

\subsection{Compression Does Not Start Before the Rise in Test Accuracy}
\label{sec:timing}

In the MLP, the compression onset follows the grokking onset in every seed (Appendix Table~\ref{tab:timing} and Figure~\ref{fig:leadlag}). All 15 seeds have $t_c>t_g$, with mean lag $6{,}367\pm3{,}248$ steps (one-sided $t=7.59$, $p=1.3\times10^{-6}$), i.e.\ $50\pm25\%$ of $t_g$. Granger tests agree: changes in test accuracy predict changes in effective rank in all 15 seeds, and the reverse holds in at most 6 (Appendix~\ref{app:tf_timing}).

In the transformer, compression is part of the transition. We locate the start $t_\text{on}$ of the second drop as the first step of the stretch leading up to $t_s$ in which the decrease is at least 20\% of its peak rate, using a centered five-point moving average that adds no delay (Appendix~\ref{app:tf_timing}). The drop starts $1{,}888\pm992$ steps before the onset, but after test accuracy has first reached 10\% in 28 of 29 seeds (median 1{,}750 steps after), at $34\pm22\%$ test accuracy. It is steepest $905\pm506$ steps after the onset (27 of 29 seeds), and 63\% of it happens after the onset. In both architectures, then, compression does not start before test accuracy begins to rise. In the MLP it starts after the transition, and in the transformer during it.

\subsection{Replication on Held-Out Tasks}
\label{sec:prereg}

We tested the timing results on seeds and tasks not used to define the measures, and report every test (Table~\ref{tab:prereg}). On the 15 held-out transformer seeds (H1), the steepest decrease came $850\pm524$ steps after the onset (14 of 15 seeds, the fifteenth at the onset, one-sided $t=6.28$, $p=1.0\times10^{-5}$), close to the $964\pm499$ steps of seeds 0--14, on which the measure had been defined. On $ab\bmod97$ and $(a{+}b)\bmod113$, the MLP's compression onset (H2) and steepest decrease (H3) followed the grokking onset in all 30 runs. The narrowing and re-widening of $d_{0.99}$ (H4, H5) and the effect of the factorization (H6) also replicated. With a Bonferroni correction over the nine $t$-tests, every test except H4 on $(a{+}b)\bmod113$ ($p=0.013$) remains significant.

\subsection{Forcing Low Rank from Initialization}
\label{sec:bottleneck}

Forcing low rank before the transition does not bring grokking forward (Table~\ref{tab:tf_caps}, Table~\ref{tab:bottleneck}). We write a weight matrix as $W=AB$ with inner dimension $r$, initialized so that the entries of $AB$ have the variance of the original initialization, and compare each rank cap with a factorized control whose inner dimension is at full rank. The control matters because the factorization alone changes the grokking time.

\paragraph{Transformer.}
The transformer's function needs only about 13 directions of $W_\text{in}$ and 8 of $W_E$, so caps above these numbers leave the solution's capacity intact. Capping $W_\text{in}$ at 32 directions leaves the onset where the factorized control has it ($14{,}350\pm5{,}730$ against $14{,}100\pm4{,}095$ steps), capping it at 24 or 16 delays it and makes it less reliable (four and three of five seeds grok), and capping it at 8, below the 13 directions the solution uses, prevents it in four of five seeds. Capping $W_E$ is worse: five of twenty capped runs grok. The four that grok at rank 24 do so at $16{,}312\pm2{,}633$ steps, against $18{,}050\pm5{,}917$ for the control. Whatever the cap, the capped matrix ends at an effective rank of 8--12, like its factorized control.

\begin{table}[tbp]
\centering
\small
\setlength{\tabcolsep}{4pt}
\caption{\textbf{Rank caps in the transformer} ($W=AB$ with inner dimension $r$, 5 seeds per arm, $t_g$ averaged over the seeds that grok, mean $\pm$ s.d.). $r=128$ is the factorized control without a rank limit.}
\label{tab:tf_caps}
\begin{tabular}{llccc}
\toprule
Matrix & $r$ & Grokked & $t_g$ & Final test accuracy (\%) \\
\midrule
None (unfactorized) & -- & 5/5 & $10{,}500\pm1{,}777$ & $100.0\pm0.0$ \\
\midrule
$W_\text{in}$ & 128 & 5/5 & $14{,}100\pm4{,}095$ & $99.6\pm0.3$ \\
 & 32 & 5/5 & $14{,}350\pm5{,}730$ & $98.4\pm1.6$ \\
 & 24 & 4/5 & $16{,}312\pm3{,}332$ & $95.5\pm7.2$ \\
 & 16 & 3/5 & $16{,}417\pm6{,}682$ & $94.5\pm7.7$ \\
 & 8 & 1/5 & $12{,}500$ & $50.6\pm39.7$ \\
\midrule
$W_E$ & 128 & 5/5 & $18{,}050\pm5{,}917$ & $100.0\pm0.0$ \\
 & 32 & 1/5 & $15{,}750$ & $31.8\pm39.1$ \\
 & 24 & 4/5 & $16{,}312\pm2{,}633$ & $83.1\pm37.4$ \\
 & 16 & 0/5 & -- & $11.9\pm9.5$ \\
 & 8 & 0/5 & -- & $13.7\pm11.8$ \\
\bottomrule
\end{tabular}
\end{table}

\paragraph{MLP.}
In the MLP the factorization alone accelerates grokking seven- to eight-fold ($t_g=1{,}667\pm58$ for $W_1=AB$ at full rank against $12{,}483\pm340$ for an unfactorized $W_1$, logged every 50 steps, and $7.2$ and $8.4$ times on the two replication tasks), whereas it slows the transformer ($14{,}100\pm4{,}095$ against $10{,}500\pm1{,}777$ steps for $W_\text{in}$, Table~\ref{tab:tf_caps}). The factorization's own effect is consistent with two known mechanisms. For gradient descent on linear networks, \citet[\S5]{arora2018acceleration} show that factorizing a matrix preconditions the updates of the product, combining an adaptive learning rate with a form of momentum, which can accelerate convergence. Weight decay on $A$ and $B$ penalizes the nuclear norm of $W$, since $\|A\|_F^2+\|B\|_F^2\geq2\|AB\|_*$. This penalty is negligible at the MLP's weight decay of $10^{-2}$ but strong at the transformer's $1.0$, consistent with the factorization speeding up the MLP and slowing the transformer. Against the factorized control, rank limits delay grokking ($2{,}450\pm100$ steps at $r=64$, $4{,}133\pm29$ at $r=32$), and models with $r\leq16$ do not grok within $10^5$ steps. Compared with the unfactorized $W_1$ alone, low rank would wrongly appear to accelerate grokking (Appendix~\ref{app:rank}). Because the MLP's solution uses most of its directions, these delays say less about compression than the transformer caps do. A nuclear-norm penalty $\lambda\|W_1\|_*$ with $\lambda=10^{-7}$ from step 0 compresses $W_1$ well before grokking (effective rank $78.5\pm3.4$ at step 11,500, against $115.9\pm0.3$) and delays grokking from $12{,}700\pm447$ to $16{,}200\pm447$ steps, and $\lambda\geq3\times10^{-7}$ prevents learning (5 seeds per $\lambda$).

\subsection{IPR Is Not a Specific Early-Warning Signal}
\label{sec:ipr}

A hidden unit of the generalizing solution reads its input as a single cosine wave over the residues. The \textbf{Fourier IPR} measures how close the units are to this. Following \citet{doshi2024togrok}, with $\hat w_k(f)$ the magnitude of the discrete Fourier transform of unit $k$'s weights, normalized to unit $\ell_2$ norm,
\begin{equation}
\text{IPR}_{\mathcal{F}}(W_1)=\frac{1}{128}\sum_{k=1}^{128}\sum_{f=0}^{p-1}\hat w_k(f)^4 .
\end{equation}
We use the weights on operand $a$ only, where \citet{doshi2024togrok} average over each unit's input and output weights. A unit tuned to a single frequency has IPR $1/2$, and unstructured weights give about $0.02$--$0.03$. For the transformer we compute it for the embedding, over the 97 tokens. \citet{doshi2024togrok} use it to separate generalizing from memorizing neurons and show that it rises during grokking. We test whether this rise can serve as a label-free early-warning signal of the onset.

IPR rises before the grokking onset, but also in every network that never groks, so it tracks the progress of training rather than an impending onset (Table~\ref{tab:ipr_sens}). In the MLP, the Fourier IPR crosses a $3\sigma$ threshold at the end of whatever baseline window is chosen and fires in all five random-label networks. In the transformer, it fires before $t_g$ in nearly every grokked seed, but also in all seven runs that do not grok (Appendix~\ref{app:ipr}).

\section{Subspace Crystallization in the Transformer}
\label{sec:transformer}

Time-resolved results are means $\pm$ s.d.\ over the 29 grokked seeds, and final-checkpoint results over the 14 grokked seeds of 0--14.

\paragraph{The transformer crystallizes during the transition.}
With a checkpoint every 250 steps and each seed aligned at its own onset, $d_\text{out}$ narrows during the rise in test accuracy (Figure~\ref{fig:tf_dense}, Table~\ref{tab:tf_dense}). Three thousand steps before the onset, at 16\% test accuracy, $d_\text{out}$ is 27--53 per matrix. At the onset it is 11--17, and 3{,}000 steps later 9--14. For $W_\text{in}$, $W_\text{out}$ and $W_U$ it is half-way down 740--960 steps before the onset, in 28 or 29 of 29 seeds, at about 60\% test accuracy. $W_E$ narrows at the onset and keeps narrowing afterwards. The narrowing coincides with the cleanup phase of \citet{nanda2023progress}. Their restricted loss, which keeps only the key-frequency components of the logits, is half-way down $1{,}966\pm647$ steps before the onset, as the generalizing circuit forms. Their excluded loss, which removes these components, is half-way up $1{,}026\pm646$ steps before the onset, as the memorizing components are removed, at the same time as $d_\text{out}$ narrows. $d_\text{out}$ adds that the cleanup reduces the directions the outputs depend on, not only those the predictions depend on, and the MLP trained with mean-squared error shows no equivalent (\S\ref{sec:crystallisation}). Transformers that never grok narrow only slowly and stay two to nine times wider (Appendix~\ref{app:dense}).

\begin{table}[tbp]
\centering
\small
\setlength{\tabcolsep}{4pt}
\caption{\textbf{The transformer crystallizes during the transition} (29 grokked seeds, each aligned at its own onset $t_g$, with checkpoints every 250 steps). $d_\text{out}$ of each matrix (test inputs, mean $\pm$ s.d.), test accuracy, and the restricted and excluded training losses of \citet{nanda2023progress} (means).}
\label{tab:tf_dense}
\begin{tabular}{rccccccc}
\toprule
Step $-\,t_g$ & $W_E$ & $W_\text{in}$ & $W_\text{out}$ & $W_U$ & Test acc.\ (\%) & Restricted & Excluded \\
\midrule
$-3{,}000$ & $27.3\pm2.8$ & $28.0\pm3.3$ & $46.9\pm7.6$ & $52.6\pm7.3$ & $15.7$ & $2.09$ & $4.8$ \\
$-2{,}000$ & $26.0\pm2.7$ & $25.7\pm3.3$ & $40.2\pm8.1$ & $46.6\pm7.9$ & $26.2$ & $1.37$ & $8.1$ \\
$-1{,}000$ & $23.0\pm3.1$ & $21.0\pm3.5$ & $26.6\pm7.4$ & $33.4\pm8.7$ & $57.7$ & $0.64$ & $14.7$ \\
$-500$ & $19.9\pm3.0$ & $18.3\pm2.9$ & $16.7\pm5.5$ & $24.2\pm8.0$ & $82.4$ & $0.26$ & $16.6$ \\
$0$ & $17.2\pm2.6$ & $15.9\pm2.3$ & $11.0\pm2.6$ & $14.0\pm4.9$ & $96.9$ & $0.12$ & $16.2$ \\
$+1{,}000$ & $12.0\pm2.2$ & $13.9\pm1.8$ & $9.2\pm1.9$ & $9.1\pm1.9$ & $99.9$ & $0.03$ & $14.0$ \\
$+3{,}000$ & $8.6\pm2.0$ & $13.8\pm1.7$ & $9.0\pm1.8$ & $8.9\pm1.7$ & $100.0$ & $0.12$ & $18.4$ \\
\bottomrule
\end{tabular}
\end{table}

\paragraph{The final solution uses few directions.}
At the final checkpoint, every analyzed matrix except $W_V$ is far below full rank (Table~\ref{tab:tf_dims}). $d_{0.99}$ is $7.7\pm1.5$ of 97 for $W_E$, $8.7\pm1.5$ of 128 for $W_\text{out}$ and $13.0\pm1.3$ of 128 for $W_\text{in}$, and the MLP post-activations need $17.2\pm3.3$ of 512 directions (3.4\%, truncating the mean-centered activation matrix). These counts are not inflated by the margin: $d_\text{out}$ is 8.1--13.0 for $W_E$, $W_\text{in}$, $W_\text{out}$ and $W_U$.

\paragraph{Runs that never grok keep high-rank matrices.}
The seven runs that memorize without generalizing have far higher-rank MLP and unembedding matrices. Counting the singular directions that carry 99\% of $\sum_i\sigma_i^2$, they keep 68--84 for $W_\text{in}$ (grokked: $18.1\pm3.8$), 61--79 for $W_\text{out}$ ($11.8\pm2.5$) and 62--73 for $W_U$ ($23.8\pm2.8$) (Table~\ref{tab:negatives}). With the comparison at matched steps in \S\ref{sec:fit} (Table~\ref{tab:matched}), this shows that low rank in these matrices is acquired at the transition, not by training to zero loss under weight decay.

\section{Why the Transformer Crystallizes Further Than the MLP}
\label{sec:comparison}

We attribute this gap to the architecture. The transformer's embedding uses about four frequencies: $4.2\pm0.7$ of them hold 90\% of its Fourier power. Each frequency requires two directions, a cosine and a sine, so this predicts about 8 directions, close to $d_{0.99}$ of $W_E$, $W_\text{out}$ and $W_U$ (7.7--8.8, Table~\ref{tab:comparison}). The directions attention writes to nearly coincide with those $W_\text{in}$ reads from (mean cosine of the principal angles $0.923$, against $0.248$ for random subspaces), and equally so in runs that never grok (Appendix~\ref{app:transformer}). The MLP block therefore receives an input already confined to a few directions, and the directions its outputs do not need are removed during the transition, together with the memorizing components (\S\ref{sec:transformer}). The MLP's first layer instead reads both one-hot operands directly, so each frequency costs it four directions, and it spreads its weights over most of the 48 frequencies (\S\ref{sec:universality}). No MLP we trained crystallizes during the transition.

Other differences between the models, i.e. loss and the weight decay do not account for the gap. The loss explains only why the cross-entropy MLP ends with fewer representative directions than the mean-squared-error MLP. A one-hot target has equal power at all 97 Fourier components, so for a network that matches its targets, omitting one frequency changes the outputs by $\sqrt{2/97}\approx14\%$, more than $d_\text{out}$ allows. CE constrains only the ordering of the logits, and the CE MLP ends at $d_\text{out}$ = 70--81, against 122--127 with MSE. Raising its weight decay to the transformer's per-step shrinkage of $10^{-3}$ speeds up grokking, from step $10{,}100$ to $2{,}300$, but does not make it crystallize: $d_\text{out}$ is about 120 just after the onset and ends at 118.4 instead of 69.6, and stronger decay prevents learning (Table~\ref{tab:ce_wd}). Only the CE MLP's $d_\text{out}$ falls late, and its Stage-4 fall in effective rank is four times as large ($37.9\%$ against $9.7\%$, Table~\ref{tab:variants}), which suggested that the late fall is Stage~4 itself (\S\ref{sec:fourstage}). Consistently, at every weight decay $d_\text{out}$ ends close to the effective rank of $W_1$ (69.6 against 71.1, 100.4 against 98.7, 118.4 against 115.7). The cross-entropy MLP's late narrowing is thus a different process from the transformer's crystallization.

\section{Discussion}
\label{sec:implications}

\paragraph{What our measurements imply for rank analyses.}
Our results suggest four practices for measuring how many directions a network uses. First, count directions with a criterion that is independent of the margin, such as $d_\text{out}$, and compare the count with that of untrained or memorizing networks: in our MLP, $d_{0.99}$ reported a narrowing that the outputs did not undergo. Second, report the training step with every count: the MLP's $d_{0.99}$ grows from 65 to 89 after generalization, so it is not a fixed property of the trained model. Third, compare any intervention that imposes low rank with a control of the same parameterization: against the unfactorized MLP, rank caps would wrongly appear to accelerate grokking. Fourth, read weight-spectrum statistics such as IPR or effective rank as signals of grokking only next to negative controls, since they also move in networks that never grok. In the transformer, a few leading singular directions of each matrix carry the solution and truncating to them keeps the test accuracy, while random truncations of equal rank do not (Table~\ref{tab:controls}, subject to the assumptions in Appendix~\ref{app:interventions}). In the MLP, one seed's hidden representation transfers to another after an orthogonal alignment fitted on its few thousand training inputs, although their weight subspaces overlap no more than random ones (\S\ref{sec:universality}).

\paragraph{Limitations.}
Our tasks are all modular arithmetic on cyclic groups, so a non-abelian group \citep{chughtai2023toy} would test the timing results more broadly. Whether language or vision models have a comparably clean sufficient subspace is a natural next question. \citet{UWS,EigneLorax} study low-dimensional weight subspaces in such models. Within our setting, the two architectures differ in more than their structure: they are also trained with different losses and for different lengths. We varied weight decay separately and found that it does not explain the difference in crystallization (Appendix~\ref{app:ce_wd}), but the cross-entropy MLPs need larger learning rates than the mean-squared-error ones to grok within $10^5$ steps (one of five seeds groks at $10^{-3}$). Finally, the Granger tests capture only temporal order, so our causal conclusions rest on the rank interventions.

\section{Conclusion}

We asked whether rank compression causes grokking (Q1) and whether the learned solution comes to lie in a few directions of each weight matrix (Q2). For Q1, compression follows the grokking onset rather than memorization: it tracks the onset across a threefold range of onset times while the training set is memorized within 250 steps, it does not start before test accuracy rises, and forcing low rank from initialization does not bring grokking forward. Compression is therefore a signature of grokking rather than its cause. For Q2, the answer depends on how directions are counted. Judged by the outputs, the transformer crystallizes during the transition, to 8--13 directions per matrix, at the same time as its memorizing components are removed. The MLP trained with mean-squared error does not: its $d_{0.99}$ narrows only because its margins grow, and its late compression tracks the optimizer's effective learning rate. Counting directions by the outputs rather than by accuracy, and comparing any imposed low rank with a control of the same parameterization, is what makes these distinctions visible.

\section*{AI Use Statement}
We used generative AI tools to retrieve and discover relevant literature and to draft and polish portions of the manuscript where appropriate. Any AI outputs were heavily edited by the authors. We did not use generative AI to formulate or refine the research hypotheses, develop the methodology, design the experiments, or interpret the results. The authors take full responsibility for the final content of this work, including its text, claims, code and data.

\bibliographystyle{plainnat}

\bibliography{references}


\appendix

\newpage
\section{Further Related Work}
\label{app:related_extended}

\paragraph{Grokking beyond modular arithmetic.}
Grokking also occurs for adversarial robustness \citep{humayun}, in multi-hop reasoning \citep{Abramov2025Grokking} and in Recursive Feature Machines \citep{mallinar2025emergence}. \citet{grocalign2025} remove the delay with spherical constraints on the residual stream. Our factorized control shows a related effect: factorizing $W_1$ at full rank shortens the delay seven- to eight-fold in the MLP (\S\ref{sec:bottleneck}).

\paragraph{Theories of the late phase.}
\citet{lyu2024dichotomy} prove that homogeneous networks with large initialization and small weight decay move from kernel-like dynamics early in training to a max-margin solution late in training, and \citet{mohamadi2023grokking} show that a margin-maximization bias explains grokking in two-layer quadratic networks on modular arithmetic, with sample complexity $\tilde{O}(p^{5/3})$ against $O(p^2)$ for kernel methods. Their setting, an exponential-type loss with weight decay, differs from our mean-squared-error MLP, which groks at the same step without weight decay (Table~\ref{tab:variants}). \citet{morwani2024feature} show that Fourier features and irreducible representations emerge from implicit margin maximization, consistent with the Fourier structure our networks learn without any spectral penalty.

\paragraph{Weight geometry and representations.}
\citet{li2018intrinsic} train networks in low-dimensional random subspaces of parameter space, and \citet{yunis2024rank} relate rank minimization, weight alignment and weight decay beyond grokking. \citet{raghu2017svcca} compare representations across networks with SVCCA, and our cross-seed results extend this comparison to grokking: the representations differ in raw coordinates and agree up to rotation. Sparse autoencoders \citep{bricken2023monosemanticity,cunningham2023sparse} decompose representations into features, a basis complementary to the singular directions used here.

\section{Experimental Details and Training Variants}
\label{app:setup}

\subsection{Data, Training and Metrics}

\paragraph{Data and training.}
The MLP uses a 35\% train split (3,293 of the 9,409 pairs) and logs metrics every 500 steps. The transformer uses a 30\% split (2,822 pairs) and logs every 250 steps. Its vocabulary comprises $p+1=98$ tokens (the residues and the equals sign), so $W_E\in\mathbb{R}^{97\times128}$ (the residue rows), $W_U\in\mathbb{R}^{128\times98}$, $W_\text{in}\in\mathbb{R}^{128\times512}$, $W_\text{out}\in\mathbb{R}^{512\times128}$, $W_Q^h,W_K^h,W_V^h\in\mathbb{R}^{128\times32}$ per head and $W_O\in\mathbb{R}^{128\times128}$ (heads stacked). The transformer's higher weight decay ($1.0$ against $10^{-2}$) follows \citet{nanda2023progress}. MLP variants use seeds 0--4, rank-constrained MLPs seeds 0--2, and transformer rank caps seeds 0--4. All experiments were run on NVIDIA RTX A5000 GPUs.

\paragraph{Training objectives.}
The MLP minimizes the mean-squared error to one-hot targets,
$\mathcal{L}_{\mathrm{MSE}} = \frac{1}{p\,|\mathcal{D}_{\mathrm{train}}|}\sum_{(a,b)}\|f(a,b)-e_{y}\|_2^2$ with $y=(a+b)\bmod p$,
and the transformer the cross-entropy over its output logits $z\in\mathbb{R}^{p+1}$,
$\mathcal{L}_{\mathrm{CE}} = -\frac{1}{|\mathcal{D}_{\mathrm{train}}|}\sum_{(a,b)}\log\big(\exp(z_{y})/\sum_{c}\exp(z_{c})\big)$.

\paragraph{Metrics.}
Stable rank is $\|W\|_F^2/\|W\|_2^2$. Hoyer sparsity is $(\sqrt{n}-\|w\|_1/\|w\|_2)/(\sqrt{n}-1)$ over the $n$ entries $w$ of $W_1$, and L0 sparsity the fraction of entries below $10^{-3}$ in magnitude. The spectral IPR $\sum_i\hat\sigma_i^4$ rises as spectral mass concentrates in a few directions. The gradient norm is the $\ell_2$ norm of the full gradient. For the transformer's Fourier IPR, each of the 128 embedding dimensions is transformed over the 97 tokens.

\begin{figure}[H]
    \centering
    \includegraphics[width=0.88\linewidth]{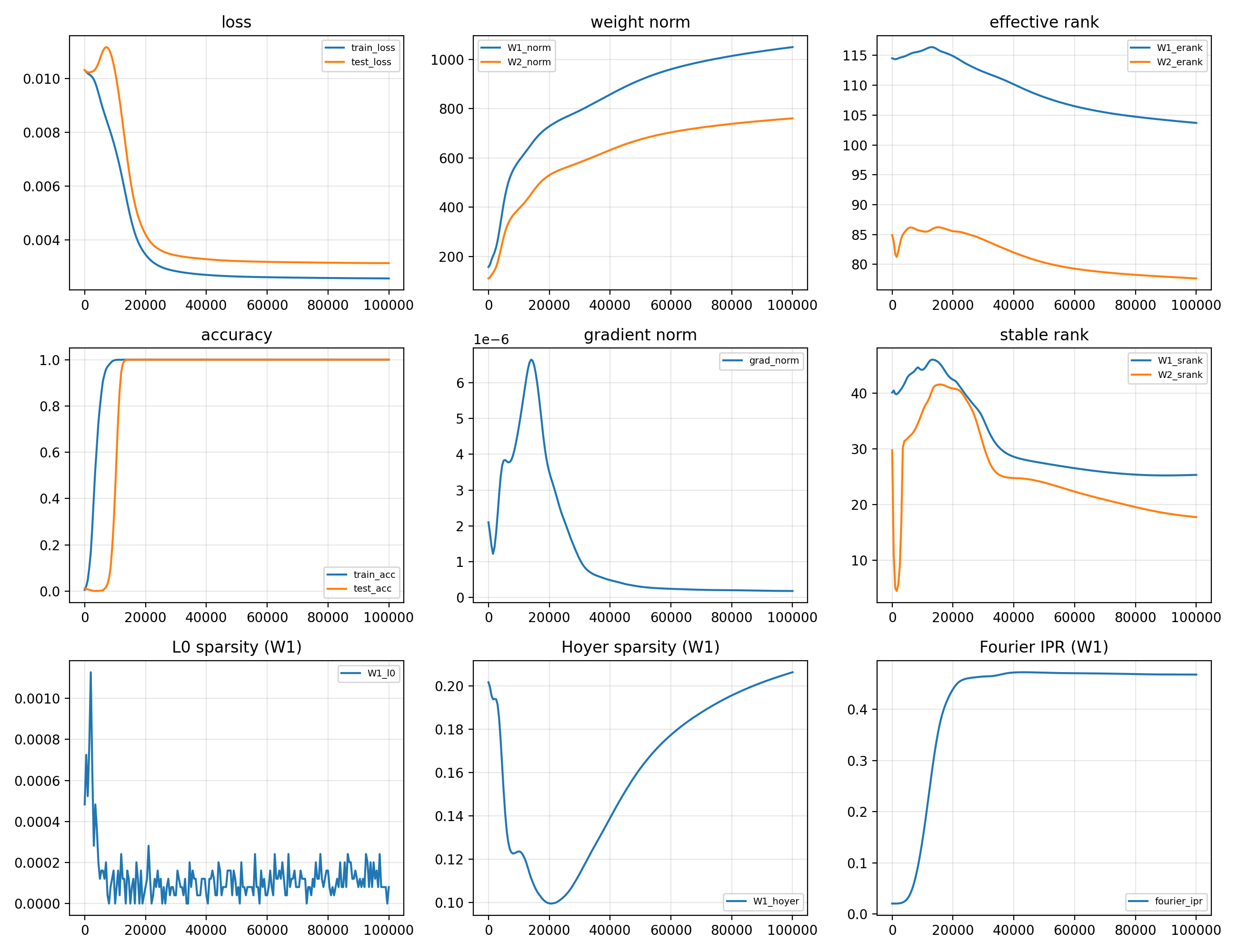}
  \caption{\textbf{Training dynamics of the MLP} (seed 0): loss, weight norms, effective rank, accuracy, gradient norm, stable rank, L0 and Hoyer sparsity of $W_1$, and Fourier IPR, against training step.}
  \label{fig:training_dynamics}
\end{figure}

\subsection{Training Variants of the MLP}
\label{app:variants}

\begin{figure}[H]
    \centering
    \begin{subfigure}[b]{0.72\linewidth}
      \includegraphics[width=\linewidth]{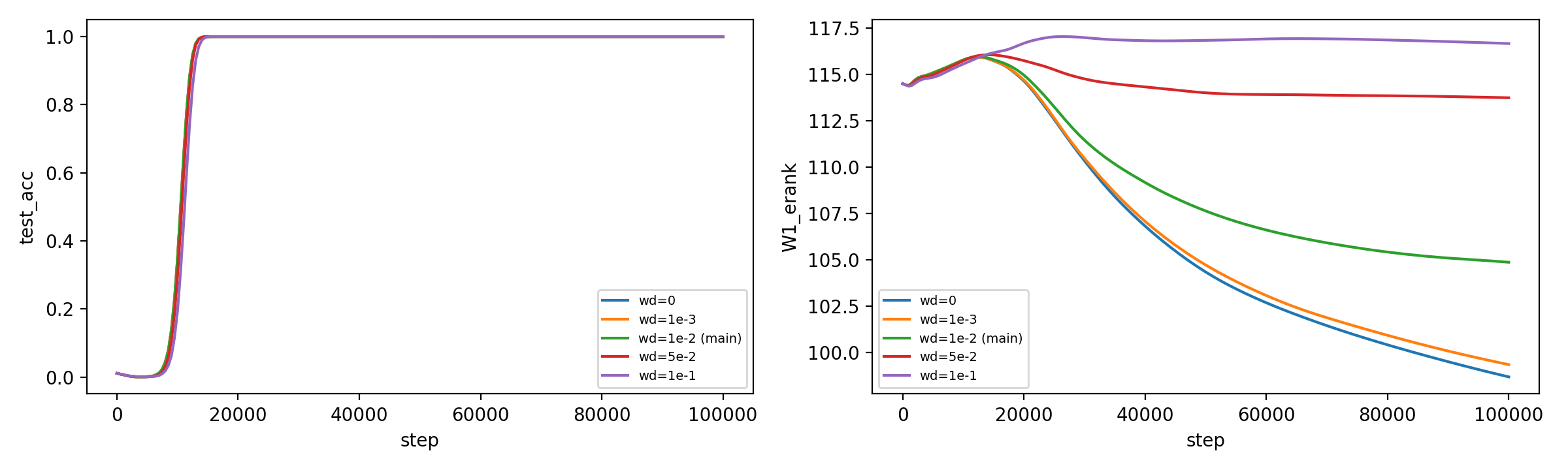}
      \caption{Weight decay.}
    \end{subfigure}
    \begin{subfigure}[b]{\linewidth}
      \includegraphics[width=\linewidth]{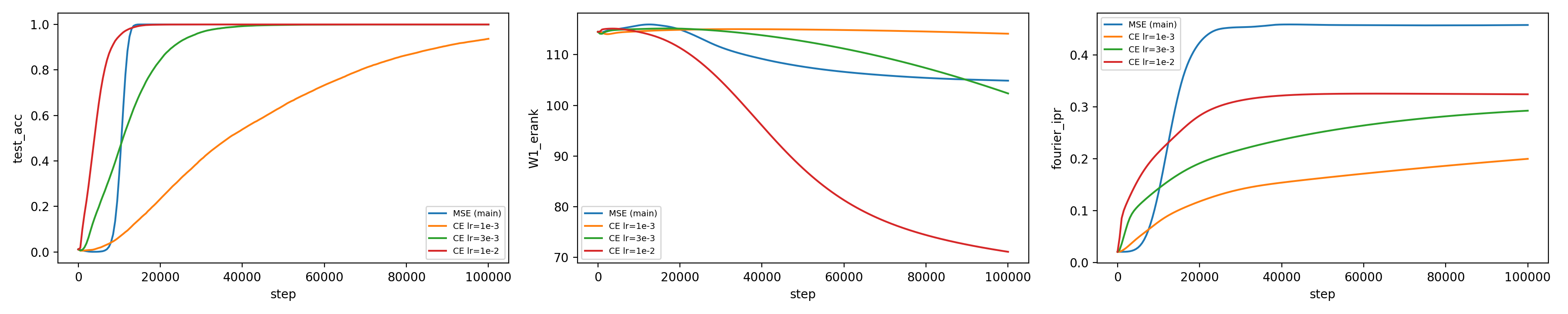}
      \caption{Loss: mean-squared error vs.\ cross-entropy at three learning rates.}
    \end{subfigure}
    \begin{subfigure}[b]{\linewidth}
      \includegraphics[width=\linewidth]{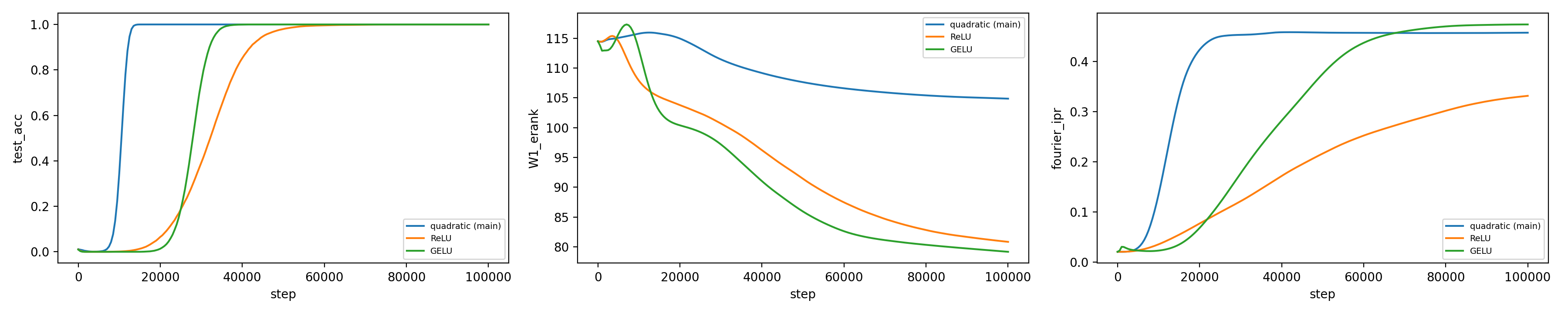}
      \caption{Activation: quadratic, ReLU and GELU.}
    \end{subfigure}
    \caption{\textbf{Stage~4 across training variants} (MLP, 5 seeds each, mean curves of test accuracy, effective rank of $W_1$ and, in (b, c), Fourier IPR).}
    \label{fig:variants}
\end{figure}

Table~\ref{tab:variants} and Figure~\ref{fig:variants} report the variants summarized in Section~\ref{sec:fourstage}.

\begin{table}[H]
\centering
\small
\setlength{\tabcolsep}{3pt}
\caption{\textbf{MLP training variants} (5 seeds each, mean $\pm$ s.d.). Main: mean-squared error, quadratic activation, weight decay (wd) $10^{-2}$. CE: cross-entropy. Grokked: test accuracy $\geq95\%$ within $10^5$ steps. Stage-4 drop: relative decrease of effective rank from its peak after $t_g$ to its final value. Fourier IPR: Section~\ref{sec:ipr}.}
\label{tab:variants}
\begin{tabular}{lccccc}
\toprule
Variant & Grokked & $t_g$ & Final ER & Stage-4 drop (\%) & Fourier IPR \\
\midrule
Main & 5/5 & $12{,}700\pm447$ & $104.9\pm1.6$ & $9.7\pm1.5$ & $0.457\pm0.009$ \\
\midrule
wd 0         & 5/5 & $12{,}700\pm447$ & $98.7\pm2.2$  & $14.9\pm1.9$ & $0.452\pm0.010$ \\
wd $10^{-3}$ & 5/5 & $12{,}700\pm447$ & $99.3\pm2.1$  & $14.4\pm1.9$ & $0.453\pm0.008$ \\
wd $0.05$    & 5/5 & $12{,}800\pm274$ & $113.7\pm0.8$ & $2.2\pm0.4$  & $0.466\pm0.006$ \\
wd $0.1$     & 5/5 & $13{,}400\pm418$ & $116.7\pm1.7$ & $0.5\pm0.4$  & $0.476\pm0.004$ \\
\midrule
CE, lr $10^{-3}$        & 1/5 & $94{,}500$ ($n=1$)   & $114.1\pm0.3$ & --           & $0.200\pm0.008$ \\
CE, lr $3\times10^{-3}$ & 5/5 & $27{,}500\pm2{,}372$ & $102.4\pm0.6$ & $10.9\pm0.5$ & $0.292\pm0.009$ \\
CE, lr $10^{-2}$        & 5/5 & $10{,}100\pm962$     & $71.1\pm2.2$  & $37.9\pm1.9$ & $0.324\pm0.011$ \\
\midrule
ReLU & 5/5 & $44{,}900\pm4{,}615$ & $80.8\pm0.8$ & $14.0\pm1.3$ & $0.332\pm0.028$ \\
GELU & 5/5 & $33{,}400\pm1{,}140$ & $79.2\pm1.8$ & $16.8\pm1.3$ & $0.474\pm0.010$ \\
\bottomrule
\end{tabular}
\\[2pt]
{\footnotesize CE at lr $10^{-3}$ reaches $93.7\pm2.0\%$ final test accuracy, and its single grokked seed has a Stage-4 drop of 0.1\%.}
\end{table}

\subsection{Weight Decay in the Cross-Entropy MLP}
\label{app:ce_wd}

Table~\ref{tab:ce_wd} trains the cross-entropy MLP at learning rate $10^{-2}$ with five weight decays $\lambda$ (5 seeds each, $10^5$ steps), including the runs of Table~\ref{tab:variants} at $\lambda=10^{-2}$. With AdamW the weights shrink by $\text{lr}\cdot\lambda$ per step, so $\lambda=0.1$ matches the transformer's per-step decay of $10^{-3}$. $d_\text{out}$ and $d_{0.99}$ are computed for $W_1$ on test inputs with centered logits, at the first saved checkpoint after $t_g$ and at the end of training. At $\lambda\geq0.3$ the weights decay to near zero, $W_1$ collapses to effective rank 1, and test accuracy stays at chance ($1.0\pm0.1\%$).

\begin{table}[H]
\centering
\small
\setlength{\tabcolsep}{4pt}
\caption{\textbf{Weight decay in the cross-entropy MLP} (learning rate $10^{-2}$, 5 seeds each, mean $\pm$ s.d.). $d_\text{out}$ and $d_{0.99}$ of $W_1$ (of 128) on test inputs with centered logits, at the first checkpoint after $t_g$ and at step $10^5$.}
\label{tab:ce_wd}
\begin{tabular}{lccccccc}
\toprule
$\lambda$ & lr$\cdot\lambda$ & Grokked & $t_g$ & $d_\text{out}$ after $t_g$ & $d_\text{out}$, end & $d_{0.99}$, end & ER$(W_1)$, end \\
\midrule
0.01 & $10^{-4}$ & 5/5 & $10{,}100\pm962$ & $119.8\pm0.4$ & $69.6\pm2.2$ & $48.4\pm2.5$ & $71.1\pm2.2$ \\
0.03 & $3\times10^{-4}$ & 5/5 & $4{,}500\pm354$ & $120.2\pm0.8$ & $100.4\pm2.3$ & $61.6\pm3.2$ & $98.7\pm2.6$ \\
0.1 & $10^{-3}$ & 5/5 & $2{,}300\pm274$ & $120.2\pm0.8$ & $118.4\pm1.7$ & $69.6\pm2.6$ & $115.7\pm1.0$ \\
0.3 & $3\times10^{-3}$ & 0/5 & -- & -- & -- & -- & $1.0\pm0.0$ \\
1.0 & $10^{-2}$ & 0/5 & -- & -- & -- & -- & $1.0\pm0.0$ \\
\bottomrule
\end{tabular}
\end{table}

\subsection{Role of the Nonlinearity}
\label{app:nonlinearity}

In the MLP, the quadratic activation turns sums of operand features into the products from which $\cos(\theta_a+\theta_b)$ is read out. ReLU and GELU do not do this exactly, yet MLPs with these activations also grok, at $t_g=44{,}900\pm4{,}615$ and $33{,}400\pm1{,}140$ against $12{,}700\pm447$, with periodic first-layer weights (Fourier IPR $0.33\pm0.03$ and $0.47\pm0.01$, against $0.46\pm0.01$) and larger Stage-4 decreases in effective rank (Table~\ref{tab:variants}). In the transformer, the ReLU of the MLP sublayer acts on inputs that attention has written into the subspace the MLP reads (alignment $0.923$, Table~\ref{tab:subspace_analysis}), and the embedding concentrates 95\% of its Fourier power on five frequencies, so the MLP sublayer operates on an already low-dimensional, low-dimensional input.

\section{Stage 4: Weight Growth and the Optimizer}
\label{app:stage4}

\subsection{Why the Sufficient Subspace Widens}
\label{app:causal_dim}

\begin{figure}[H]
  \centering
  \includegraphics[width=\linewidth]{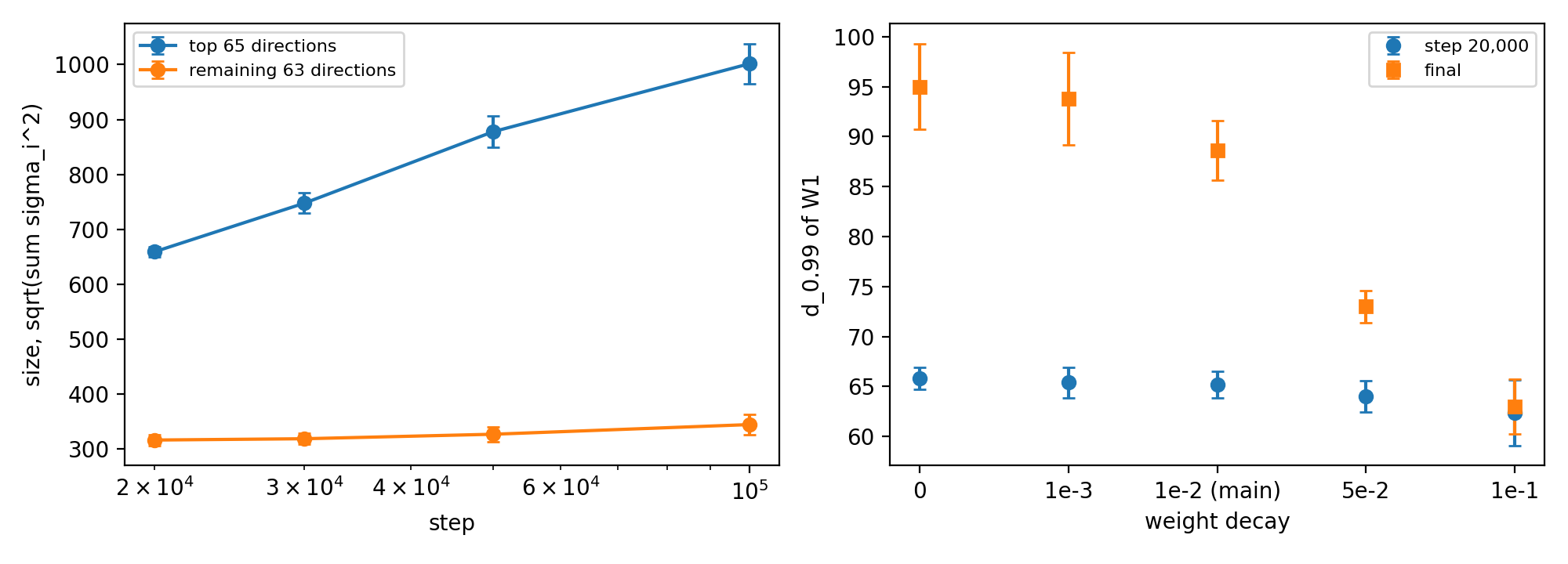}
  \caption{\textbf{Stage-4 widening.} Left: size of the top 65 and the remaining 63 singular directions of $W_1$ over training (15 seeds). Right: $d_{0.99}$ at step 20,000 and at the end of training for each weight decay (5 seeds).}
  \label{fig:widening}
\end{figure}

Table~\ref{tab:widening} and Figure~\ref{fig:widening} decompose the widening of the sufficient subspace for the main runs, and Table~\ref{tab:widening_wd} repeats the measurement across weight decays. The top 65 directions of $W_1$ grow by half while the remaining 63 keep their size. The output change caused by removing the remaining directions, dominated by the cross term $2h_\text{top}h_\text{rest}$, roughly doubles while the margin grows by 30\%, so truncation to the top 65 directions loses accuracy and $d_{0.99}$ widens. Stronger weight decay limits the growth of the top directions and, with it, both the fall in effective rank and the widening.

\begin{table}[H]
\centering
\small
\setlength{\tabcolsep}{3pt}
\caption{\textbf{Why the sufficient subspace widens in Stage~4} (MLP, 15 seeds). $K=65$ is the median $d_{0.99}$ at step 20,000. Size: $(\sum_i\sigma_i^2)^{1/2}$ over the top 65 or the remaining 63 singular directions of $W_1$. Output change: median $\ell_2$ change of the 97 outputs when the directions beyond the top 65 are removed, split into the two terms of $(h_\text{top}+h_\text{rest})^2-h_\text{top}^2$.}
\label{tab:widening}
\begin{tabular}{lcccc}
\toprule
Step & 20,000 & 30,000 & 50,000 & 100,000 \\
\midrule
$d_{0.99}$ & $64.9\pm1.7$ & $69.3\pm1.7$ & $81.5\pm2.4$ & $89.1\pm2.3$ \\
ER$(W_1)$ & $114.4\pm1.1$ & $110.9\pm1.5$ & $107.0\pm1.9$ & $104.0\pm2.1$ \\
$\|W_1\|_F$ & $731\pm7$ & $813\pm14$ & $937\pm26$ & $1059\pm35$ \\
$\|W_2\|_F$ & $531\pm5$ & $597\pm11$ & $685\pm21$ & $763\pm27$ \\
$\sigma_1(W_1)$ & $111.9\pm3.4$ & $132.2\pm3.0$ & $171.1\pm10.4$ & $217.8\pm21.6$ \\
Size of top 65 & $659\pm10$ & $748\pm18$ & $878\pm29$ & $1002\pm37$ \\
Size of remaining 63 & $316\pm10$ & $318\pm10$ & $327\pm13$ & $344\pm18$ \\
Test acc., top-65 truncation (\%) & $99.1\pm0.4$ & $97.7\pm0.9$ & $84.5\pm5.5$ & $57.7\pm9.5$ \\
Median correct-class output & $0.604\pm0.017$ & $0.696\pm0.010$ & $0.722\pm0.008$ & $0.732\pm0.007$ \\
Median margin & $0.470\pm0.019$ & $0.570\pm0.012$ & $0.599\pm0.009$ & $0.611\pm0.008$ \\
Output change, $2h_\text{top}h_\text{rest}$ & $0.611\pm0.015$ & $0.699\pm0.014$ & $0.860\pm0.063$ & $1.153\pm0.128$ \\
Output change, $h_\text{rest}^2$ & $0.254\pm0.019$ & $0.336\pm0.022$ & $0.445\pm0.029$ & $0.533\pm0.047$ \\
\bottomrule
\end{tabular}
\end{table}

\begin{table}[H]
\centering
\small
\setlength{\tabcolsep}{4pt}
\caption{\textbf{Weight decay controls the widening} (5 seeds each, step 20,000 $\to$ step $10^5$, means). All runs reach 100\% test accuracy.}
\label{tab:widening_wd}
\begin{tabular}{lccccc}
\toprule
Weight decay & $d_{0.99}$ & ER$(W_1)$ & $\|W_1\|_F$ & $\sigma_1(W_1)$ & Size of remaining 63 \\
\midrule
0                 & $65.8\to95.0$ & $114.6\to98.7$  & $731\to1{,}252$ & $114\to251$ & $320\to358$ \\
$10^{-3}$         & $65.4\to93.8$ & $114.7\to99.3$  & $731\to1{,}220$ & $113\to249$ & $320\to354$ \\
$10^{-2}$ (main)  & $65.2\to88.6$ & $115.0\to104.9$ & $727\to1{,}043$ & $111\to205$ & $320\to347$ \\
$5\times10^{-2}$  & $64.0\to73.0$ & $115.7\to113.7$ & $710\to825$     & $101\to151$ & $321\to342$ \\
$10^{-1}$         & $62.4\to63.0$ & $116.7\to116.7$ & $678\to748$     & $92\to108$  & $311\to346$ \\
\bottomrule
\end{tabular}
\end{table}

\subsection{Gradient and Adam State}
\label{app:adam}

We continue each main seed for 1{,}000 AdamW steps from the checkpoints at steps 20{,}000, 50{,}000 and $10^5$ to rebuild the optimizer state, and then measure the full-batch gradient $g$ and Adam's bias-corrected moments $\hat m,\hat v$ (Table~\ref{tab:adam_state}). The gradient is small and becomes increasingly radial: its cosine with $\theta$ goes from $-0.19$ to $-0.89$, so $-g$ points outward and the loss would decrease if the weights grew. This radial component equals the Euler expression $\langle g,\theta\rangle=3\cdot\tfrac{2}{p|\mathcal{D}|}\sum(f-y)f$ for this 3-homogeneous model to all printed digits. Almost every coordinate has $\sqrt{\hat v}<\epsilon=10^{-8}$, so the update $\hat m/(\sqrt{\hat v}+\epsilon)$ is close to $\hat m/\epsilon$: AdamW acts as gradient descent with momentum and learning rate $\text{lr}/\epsilon=10^5$, not as sign descent (the mean of $|\hat m|/(\sqrt{\hat v}+\epsilon)$ is 0.07--0.21, against 1 for sign descent). The radial part of this update exceeds the decay pull at step 20{,}000 and approaches it by the end, as the norm approaches an equilibrium. The gradient also concentrates in the top 65 directions of $W_1$, the directions that grow.

\begin{table}[H]
\centering
\small
\setlength{\tabcolsep}{4pt}
\caption{\textbf{Gradient and Adam state during Stage~4} (MLP, 15 seeds, mean $\pm$ s.d.). Radial push of the update and decay pull $\lambda\|\theta\|^2$ are per unit learning rate.}
\label{tab:adam_state}
\begin{tabular}{lccc}
\toprule
 & Step 20{,}000 & Step 50{,}000 & Step $10^5$ \\
\midrule
$\|g\|$ & $3.4\times10^{-6}$ & $3.2\times10^{-7}$ & $1.9\times10^{-7}$ \\
$\langle g,\theta\rangle$ (= Euler expression) & $-6.0\times10^{-4}$ & $-2.5\times10^{-4}$ & $-2.2\times10^{-4}$ \\
$\cos(g,\theta)$ & $-0.19\pm0.01$ & $-0.67\pm0.07$ & $-0.89\pm0.06$ \\
Share of coordinates with $\sqrt{\hat v}<\epsilon$ & $0.870\pm0.017$ & $0.998\pm0.001$ & $0.9998\pm0.0002$ \\
Median $\sqrt{\hat v}$ & $2.1\times10^{-9}$ & $7.3\times10^{-10}$ & $6.1\times10^{-10}$ \\
Mean $|\hat m|/(\sqrt{\hat v}+\epsilon)$ (1 = sign descent) & $0.21\pm0.01$ & $0.08\pm0.01$ & $0.07\pm0.00$ \\
Radial push of the Adam update & $(2.0\pm0.1)\times10^{4}$ & $(2.0\pm0.2)\times10^{4}$ & $(1.9\pm0.1)\times10^{4}$ \\
Decay pull $\lambda\|\theta\|^2$ & $(0.84\pm0.02)\times10^{4}$ & $(1.36\pm0.08)\times10^{4}$ & $(1.71\pm0.12)\times10^{4}$ \\
Share of $\|g_{W_1}\|^2$ in the top 65 directions & $0.43\pm0.03$ & $0.79\pm0.09$ & $0.87\pm0.03$ \\
\bottomrule
\end{tabular}
\end{table}

\subsection{Optimizer Swap}

Table~\ref{tab:optimizers_full} continues main seeds 0--2 from step 20{,}000 to $10^5$ with each optimizer. The weight decay is set so that the per-step shrinkage matches AdamW's $\text{lr}\cdot\lambda=10^{-5}$ ($\lambda=10^{-5}/\text{lr}$ for gradient descent, $10^{-6}/\text{lr}$ for momentum 0.9, whose steady-state step is $\text{lr}/(1-0.9)$). Runs with the same effective learning rate agree closely: SGD with momentum at lr 10 reproduces gradient descent at lr 100 to the printed digits, and AdamW with $\epsilon=10^{-6}$ reproduces gradient descent at lr $10^3$. Stage~4, the fall of effective rank with the growth of $\sigma_1$ and the widening of $d_{0.99}$, appears at effective learning rates of $10^5$ and above and strengthens as $\epsilon$ decreases. Test accuracy stays at 100\% in every run.

\begin{table}[H]
\centering
\footnotesize
\setlength{\tabcolsep}{3pt}
\caption{\textbf{Continuations from step 20{,}000 to $10^5$} (main seeds 0--2, means). The effective learning rate is lr for gradient descent, lr$/\epsilon$ for AdamW and lr$/(1-0.9)$ for SGD with momentum.}
\label{tab:optimizers_full}
\begin{tabular}{llccccc}
\toprule
Optimizer & Effective lr & $\|W_1\|_F$ & $\|W_2\|_F$ & $\sigma_1(W_1)$ & ER$(W_1)$ & $d_{0.99}$ \\
\midrule
At step 20{,}000 & -- & 726 & 529 & 111 & 115.1 & 65 \\
\midrule
AdamW, $\epsilon=10^{-10}$ & $10^7$ & 1{,}793 & 718 & 346 & 95.7 & 99 \\
AdamW, $\epsilon=10^{-9}$ & $10^6$ & 1{,}582 & 874 & 318 & 94.7 & 98 \\
AdamW, $\epsilon=10^{-8}$ (main) & $10^5$ & 1{,}040 & 748 & 203 & 104.8 & 89 \\
AdamW, $\epsilon=10^{-7}$ & $10^4$ & 775 & 551 & 120 & 114.8 & 70 \\
AdamW, $\epsilon=10^{-6}$ & $10^3$ & 658 & 469 & 85 & 117.8 & 65 \\
AdamW, $\epsilon=10^{-4}$ & $10$ & 331 & 241 & 51 & 115.3 & 65 \\
AdamW, $\epsilon=10^{-8}$, no weight decay & $10^5$ & 1{,}257 & 917 & 261 & 97.8 & 94 \\
\midrule
SGD, lr $10^{-3}$ (same decay) & $10^{-3}$ & 326 & 238 & 50 & 115.1 & 65 \\
Gradient descent & $10^2$ & 384 & 277 & 56 & 116.7 & 65 \\
Gradient descent & $10^3$ & 662 & 472 & 85 & 117.8 & 65 \\
Gradient descent & $10^4$ & 783 & 557 & 126 & 114.6 & 72 \\
Gradient descent & $10^5$ & 833 & 913 & 148 & 108.3 & 87 \\
Gradient descent & $10^6$ & \multicolumn{5}{c}{diverges} \\
\midrule
SGD + momentum, lr 10 & $10^2$ & 384 & 277 & 56 & 116.7 & 65 \\
SGD + momentum, lr $10^2$ & $10^3$ & 662 & 472 & 85 & 117.8 & 65 \\
SGD + momentum, lr $10^3$ & $10^4$ & 783 & 557 & 126 & 114.6 & 72 \\
SGD + momentum, lr $10^4$ & $10^5$ & 1{,}093 & 775 & 216 & 103.8 & 91 \\
SGD + momentum, lr $10^5$ & $10^6$ & 881 & 1{,}577 & 174 & 101.1 & 93 \\
\bottomrule
\end{tabular}
\end{table}

\section{The Sufficient Subspace, Output Counts and Cross-Seed Equivalence}
\label{app:subspace}

\subsection{The Sufficient Subspace over Training}

\begin{figure}[H]
  \centering
  \begin{subfigure}[b]{0.48\linewidth}
    \includegraphics[width=\linewidth]{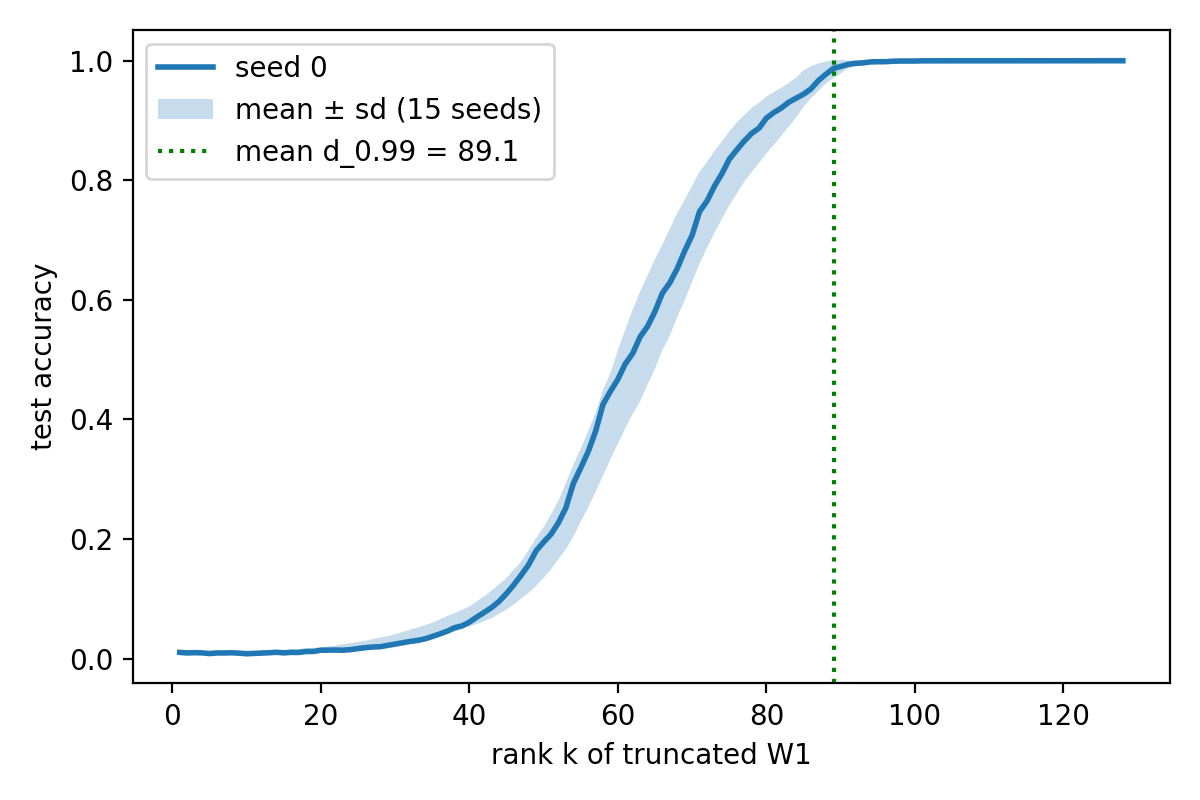}
    \caption{Test accuracy of $W_1$ truncated to its top $k$ directions (step $10^5$).}
    \label{fig:causal_dim}
  \end{subfigure}\hfill
  \begin{subfigure}[b]{0.48\linewidth}
    \includegraphics[width=\linewidth]{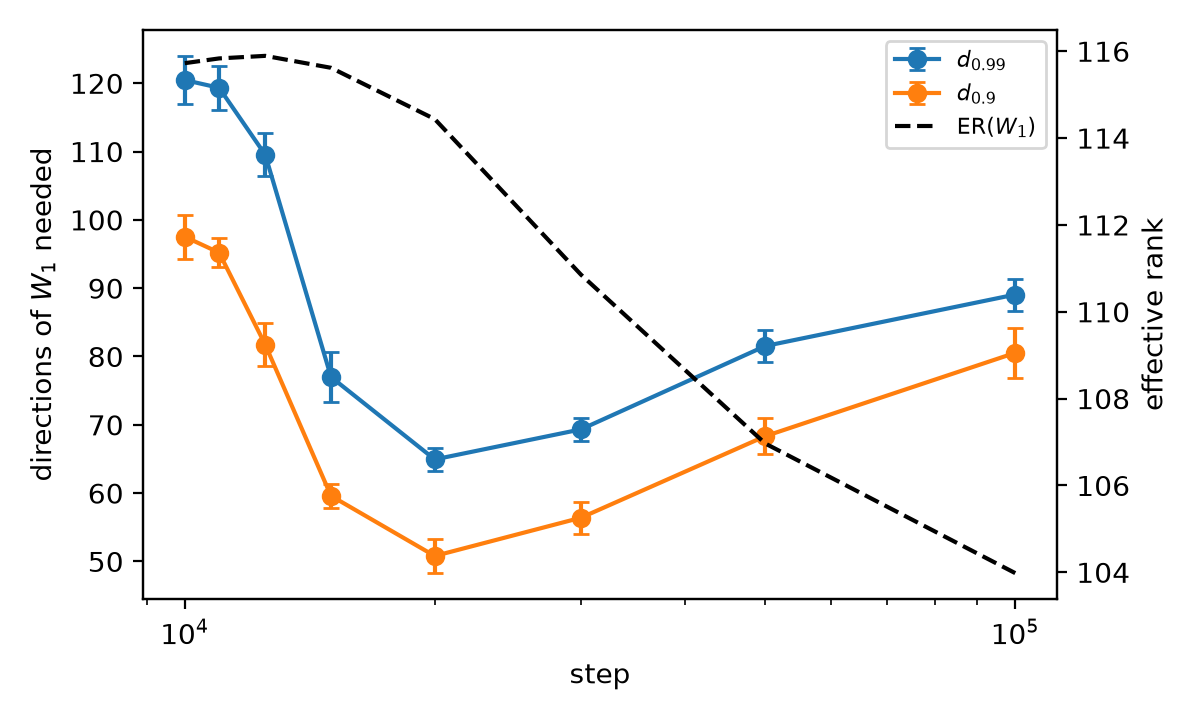}
    \caption{$d_{0.9}$ and $d_{0.99}$ (left axis) and effective rank (right axis) over training.}
    \label{fig:dc_time}
  \end{subfigure}
  \caption{\textbf{The sufficient subspace of the MLP} (15 seeds). (a) Seed 0 and mean $\pm$ s.d. (b) $d_{0.9}$ and $d_{0.99}$ narrow sharply after the onset and widen again during Stage~4, while effective rank falls. $d_\text{out}$ does not narrow (Table~\ref{tab:margin_free}).}
  \label{fig:subspace_top}
\end{figure}

Table~\ref{tab:causal_dim} reports $d_\tau$ of $W_1$ at four thresholds at the end of training, which ranges from 80 to 93 directions for thresholds from 90\% to 99.9\%. Table~\ref{tab:dc_time} and Figure~\ref{fig:subspace_top} show its evolution over training. Before step 15,000, $d_\tau$ is relative to each checkpoint's own test accuracy, which is below 100\%. The number of frequencies holding 90\% of the Fourier power of $W_1$ falls from 38.2 to 33.7 during Stage~4.

\begin{table}[H]
\centering
\small
\caption{$d_\tau$ of $W_1$ at four thresholds (step $10^5$, 15 seeds).}
\label{tab:causal_dim}
\begin{tabular}{lc}
\toprule
Threshold $\tau$ (fraction of clean accuracy) & $d_\tau$ (mean $\pm$ s.d.) \\
\midrule
90\%   & $80.5\pm3.7$ \\
95\%   & $84.5\pm3.2$ \\
99\%   & $89.1\pm2.3$ \\
99.9\% & $93.1\pm2.7$ \\
\bottomrule
\end{tabular}
\end{table}

\begin{table}[H]
\centering
\small
\caption{$d_{0.9}$, $d_{0.99}$, effective rank and frequency usage of $W_1$ over training (15 seeds).}
\label{tab:dc_time}
\begin{tabular}{rccccc}
\toprule
Step & Test acc.\ (\%) & $d_{0.9}$ & $d_{0.99}$ & ER$(W_1)$ & Freq.\ for 90\% power \\
\midrule
10,000  & $37.0\pm8.0$  & $97.5\pm3.2$ & $120.5\pm3.5$ & $115.7\pm0.4$ & $40.8\pm0.4$ \\
11,000  & $67.3\pm8.9$  & $95.2\pm2.1$ & $119.3\pm3.2$ & $115.8\pm0.4$ & $40.3\pm0.7$ \\
12,500  & $95.7\pm2.7$  & $81.7\pm3.2$ & $109.5\pm3.2$ & $115.9\pm0.6$ & $39.7\pm0.5$ \\
15,000  & $100.0\pm0.0$ & $59.6\pm1.8$ & $77.0\pm3.6$  & $115.6\pm0.8$ & $38.5\pm0.6$ \\
20,000  & $100.0\pm0.0$ & $50.8\pm2.5$ & $64.9\pm1.7$  & $114.4\pm1.1$ & $38.2\pm1.1$ \\
30,000  & $100.0\pm0.0$ & $56.4\pm2.4$ & $69.3\pm1.7$  & $110.9\pm1.5$ & $38.3\pm0.9$ \\
50,000  & $100.0\pm0.0$ & $68.3\pm2.6$ & $81.5\pm2.4$  & $107.0\pm1.9$ & $36.1\pm1.4$ \\
100,000 & $100.0\pm0.0$ & $80.5\pm3.7$ & $89.1\pm2.3$  & $104.0\pm2.1$ & $33.7\pm1.9$ \\
\bottomrule
\end{tabular}
\end{table}

\subsection{Temporal Stability}
\label{app:stability}

We compute the mean cosine of the principal angles between the top-89 left and right singular subspaces of $W_1$ at step $s$ and at the final checkpoint ($k=89$, the median $d_{0.99}$), across all 15 seeds (Table~\ref{tab:stability}). At initialization the overlap is at the random level. At $t_g$ it is only modestly above it ($0.837$ against $0.773$ in the hidden space), and it keeps rising for tens of thousands of steps. The final sufficient subspace is therefore assembled mostly during Stages~3 and~4, not at the accuracy transition.

\begin{table}[H]
\centering
\small
\caption{Overlap of the top-89 singular subspaces of $W_1$ at step $s$ with those at step $10^5$ (15 seeds). Hidden space: left singular vectors (the sufficient subspace $\mathcal{C}$). Input space: right singular vectors.}
\label{tab:stability}
\begin{tabular}{rcc}
\toprule
Step $s$ & Hidden space ($\mathcal{C}$) & Input space \\
\midrule
0       & $0.775\pm0.003$ & $0.608\pm0.003$ \\
5,000   & $0.782\pm0.003$ & $0.623\pm0.003$ \\
10,000  & $0.807\pm0.005$ & $0.677\pm0.007$ \\
12,500  & $0.837\pm0.007$ & $0.723\pm0.010$ \\
15,000  & $0.875\pm0.010$ & $0.773\pm0.012$ \\
20,000  & $0.925\pm0.013$ & $0.843\pm0.013$ \\
30,000  & $0.961\pm0.013$ & $0.910\pm0.011$ \\
50,000  & $0.983\pm0.012$ & $0.967\pm0.013$ \\
\midrule
Random  & $0.773\pm0.003$ & $0.606\pm0.003$ \\
\bottomrule
\end{tabular}
\end{table}

\subsection{Truncation Controls and Complement Interventions}
\label{app:interventions}

\begin{table}[H]
\centering
\small
\caption{\textbf{Top-$k$, bottom-$k$ and random-$k$ truncations} at equal rank $k=d_{0.99}$ (per seed). Test accuracy in \%, mean $\pm$ s.d.\ over 15 MLP seeds at step $10^5$ and 14 grokked transformer seeds. Random controls average 10 draws. Chance is 1.03\%.}
\label{tab:controls}
\setlength{\tabcolsep}{4pt}
\begin{tabular}{lccccc}
\toprule
Matrix & $k=d_{0.99}$ & Top-$k$ & Bottom-$k$ & Random subset & Random subspace \\
\midrule
MLP $W_1$                  & $89.1\pm2.3$ / 128 & $99.26\pm0.22$ & $28.26\pm6.18^{*}$ & $31.26\pm4.37$ & $22.74\pm3.93$ \\
Transformer $W_\text{in}$  & $13.0\pm1.3$ / 128 & $99.80\pm0.28$ & $5.50\pm1.46^{\ddagger}$ & $3.50\pm0.54$ & $12.49\pm2.99$ \\
Transformer $W_\text{out}$ & $8.7\pm1.5$ / 128  & $99.95\pm0.07$ & $5.50\pm1.46^{\ddagger}$ & $4.42\pm0.95$ & $8.00\pm1.93$ \\
Transformer $W_E$          & $7.7\pm1.5$ / 97   & $99.74\pm0.35$ & $1.01\pm0.06$ & $1.33\pm0.23$ & $1.31\pm0.13$ \\
Transformer $W_U$          & $8.8\pm1.4$ / 98   & $99.86\pm0.27$ & $1.07\pm0.32$ & $1.80\pm0.65$ & $6.02\pm1.92$ \\
\bottomrule
\end{tabular}
\\[2pt]
{\footnotesize $^{*}$Top-$k$ and bottom-$k$ share about 50 directions when $k>64$. $^{\ddagger}$Keeping only the bottom directions of $W_\text{in}$ or of $W_\text{out}$ effectively removes the MLP's contribution, hence the identical values.}
\end{table}

\paragraph{Complement interventions.}
We intervene on the MLP's pre-activations $h=W_1x/\sqrt{194}$, with $\mathcal{C}$ the sufficient subspace of each seed at step $10^5$. Keeping only the $\mathcal{C}$-component gives $99.26\pm0.22\%$ test accuracy and keeping only the $\mathcal{C}^\perp$-component $0.15\pm0.14\%$. Adding Gaussian noise of norm $s\|h\|$ (per input) inside $\mathcal{C}$ or inside $\mathcal{C}^\perp$ gives $99.97\pm0.05\%$ and $99.95\pm0.07\%$ for $s=0.1$, $92.42\pm4.22\%$ and $70.97\pm9.69\%$ for $s=0.3$, $10.11\pm2.06\%$ and $7.43\pm1.32\%$ for $s=1$, and $1.54\pm0.10\%$ and $1.61\pm0.09\%$ for $s=3$. As a check of the pipeline, patching the clean hidden representation into a run on noise-corrupted inputs restores the clean output (100.00\% in the MLP at corruption level 0.3, against 1.59\% unpatched, and 99.92\% in the transformer, against 18.52\%), as it must in a network with one hidden layer.

\paragraph{Assumptions of the SVD probe.}
Equating high-variance singular directions with functional relevance assumes that those directions carry the computation rather than optimizer noise, representation drift or dense isotropic features. It also assumes linear, orthogonal structure, which can obscure sparse features stored in non-orthogonal superposition or on nonlinear manifolds. The controls of Table~\ref{tab:controls} test the first assumption in our setting: the high-variance singular directions carry the computation, while random subspaces of equal rank do not. In the MLP the distinction is weaker, because the sufficient subspace spans most of the directions.

\subsection{Output Counts under Other Criteria}
\label{app:baselines}

Table~\ref{tab:baselines} gives the directions of $W_1$ needed under four output-based criteria: outputs within 10\% ($d_\text{out}$) and 5\%, per-input centered outputs within 10\%, and per-input margins within 10\%. Centering changes nothing for the MSE models. The margin criterion lies between accuracy and the full outputs: it narrows from 124 to 113 directions between steps 11{,}000 and 20{,}000 and widens again to 121, much less than $d_{0.99}$ (119 to 65 to 89). For the cross-entropy MLPs, every output-based criterion narrows only after the onset.

\begin{table}[H]
\centering
\footnotesize
\setlength{\tabcolsep}{3pt}
\caption{\textbf{Output counts in the MLP} (directions of $W_1$ out of 128, mean over seeds, test inputs unless noted). $d_\text{out}$ and Out05: outputs within 10\% and 5\% (centered for CE). Cen10: per-input centered outputs within 10\%. Marg10: per-input margins within 10\%.}
\label{tab:baselines}
\begin{tabular}{llcccccc}
\toprule
Model & Checkpoint & $d_{0.99}$ & $d_\text{out}$ & Out05 & Cen10 & Marg10 & Median margin \\
\midrule
MSE, main & step 0 & -- & 116.4 & 123.8 & 116.3 & 102.2 & $-0.004$ \\
 & step 11{,}000 & 119.3 & 122.4 & 126.6 & 122.7 & 123.8 & 0.025 \\
 & step 20{,}000 & 64.9 & 122.3 & 125.7 & 122.4 & 113.1 & 0.470 \\
 & end & 89.1 & 127.3 & 127.9 & 127.3 & 121.4 & 0.611 \\
 & end, training inputs & 88.0 & 127.3 & 127.9 & 127.5 & 121.1 & 0.649 \\
MSE, random labels & end, training inputs & -- & 123.0 & 127.0 & 123.2 & 116.6 & 0.110 \\
 & end, test inputs & -- & 121.4 & 126.0 & 121.6 & 108.0 & $-0.229$ \\
MSE, $ab\bmod97$ & last before $t_g$ & 115.9 & 122.1 & 126.3 & 122.3 & 117.5 & 0.052 \\
 & first after $t_g$ & 97.5 & 122.5 & 126.3 & 122.5 & 112.7 & 0.155 \\
 & step 20{,}000 & 64.7 & 121.7 & 125.1 & 121.7 & 112.4 & 0.481 \\
 & end & 89.8 & 126.2 & 126.9 & 126.2 & 120.7 & 0.653 \\
MSE, $(a{+}b)\bmod113$ & last before $t_g$ & 117.7 & 123.9 & 127.1 & 124.2 & 123.1 & 0.048 \\
 & first after $t_g$ & 79.4 & 124.3 & 127.1 & 124.4 & 115.3 & 0.232 \\
 & step 20{,}000 & 70.9 & 124.3 & 126.9 & 124.3 & 114.1 & 0.275 \\
 & end & 85.5 & 127.4 & 127.9 & 127.4 & 121.5 & 0.501 \\
\midrule
CE, lr $3\times10^{-3}$ & last before $t_g$ & 112.8 & 121.4 & 125.8 & 121.4 & 120.8 & 3.78 \\
 & first after $t_g$ & 91.8 & 119.0 & 125.4 & 119.0 & 112.6 & 5.86 \\
 & end & 50.6 & 81.0 & 99.8 & 81.0 & 68.0 & 9.43 \\
CE, lr $10^{-2}$ & last before $t_g$ & 104.6 & 120.2 & 125.8 & 120.2 & 117.8 & 4.22 \\
 & first after $t_g$ & 98.6 & 119.8 & 125.8 & 119.8 & 115.6 & 4.86 \\
 & end & 48.4 & 69.6 & 71.8 & 69.6 & 65.8 & 10.20 \\
\bottomrule
\end{tabular}
\end{table}

\subsection{Cross-Seed Equivalence}
\label{app:transplant}

For a source model $A$ and a receiving model $B$, we compute $A$'s post-activations $Z_A$ on $A$'s test inputs and decode them with $B$'s readout $W_2^B$, either directly or after the orthogonal map $R^\star=\arg\min_{R^\top R=I}\|Z_A^{\mathrm{tr}}R-Z_B^{\mathrm{tr}}\|_F$ fitted on $A$'s training inputs \citep{schonemann1966generalized}, and measure accuracy against the true labels $(a+b)\bmod97$ (Table~\ref{tab:cka}). Rows with random-label models test specificity in one direction only: a readout trained on random labels cannot decode the true labels from any representation, so only transplants into a grokked readout are informative. The step-10{,}000 checkpoint of each seed (at least 99\% training and $37.0\pm8.0\%$ test accuracy) already partly generalizes, so its rows do not isolate a memorizing representation.

\begin{table}[H]
\centering
\small
\caption{\textbf{Cross-seed transplants and linear CKA} (MLP, final checkpoints unless noted, mean $\pm$ s.d.). Transplant: test accuracy of model $B$'s readout applied to model $A$'s post-activations after the map (chance 1.03\%). CKA on all $p^2$ inputs.}
\label{tab:cka}
\begin{tabular}{lcc}
\toprule
Model $A$ $\to$ grokked model $B$ & Map & Test accuracy (\%) \\
\midrule
Grokked, other seed & fitted orthogonal & $95.82\pm2.38$ \\
Grokked, other seed & none & $1.03\pm0.30$ \\
Grokked, other seed & random orthogonal & $0.94\pm0.27$ \\
Random initialization & fitted orthogonal & $0.61\pm0.12$ \\
Random-label network & fitted orthogonal & $0.68\pm0.07$ \\
Grokked on $ab\bmod97$ (other task) & fitted orthogonal & $1.42\pm0.09$ \\
Same seed at step 10{,}000 & none & $36.18\pm6.40$ \\
Same seed at step 10{,}000 & fitted orthogonal & $25.09\pm5.77$ \\
\midrule
Linear CKA between & & \\
\quad grokked seeds & & $0.096\pm0.019$ \\
\quad a grokked network and its initialization & & $0.089\pm0.010$ \\
\quad two initializations & & $0.142\pm0.002$ \\
\bottomrule
\end{tabular}
\end{table}

\begin{figure}[H]
    \centering
    \includegraphics[width=\linewidth]{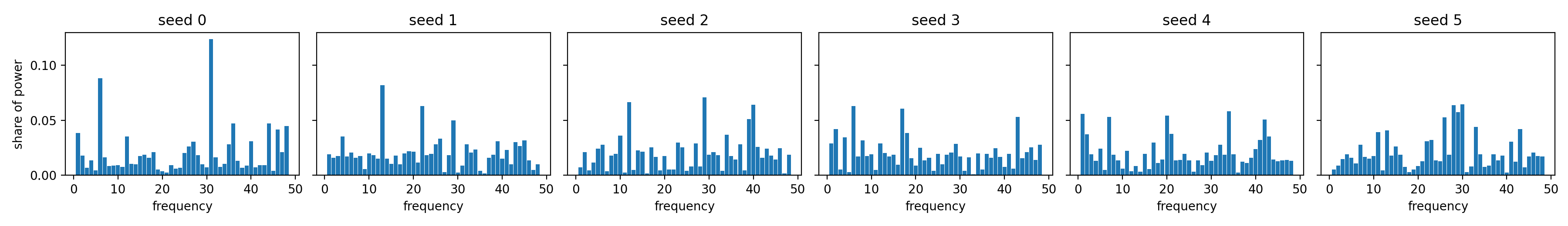}
  \caption{\textbf{Fourier power spectra of $W_1$ for six seeds} (share of power per frequency, $f$ and $p-f$ combined).}
  \label{fig:fourier_seeds}
\end{figure}

\paragraph{Subspace overlap.}
\label{app:overlap}
Figure~\ref{fig:fourier_seeds} shows the Fourier power spectrum of $W_1$ for six seeds. Overlaps are mean cosines of the principal angles between the top-60 weight subspaces of $W_1$ (in the 194-dimensional input space), and between 10-dimensional Fourier subspaces built from each seed's five strongest frequencies (a cosine and a sine for each, with $f$ and $p-f$ treated as one frequency), over all 105 seed pairs, with random baselines simulated for the same dimensions. A few individual directions are shared: per seed pair, $5.78\pm1.97$ of the top-60 directions have a counterpart with cosine above $0.9$, against 1.38 expected for random subspaces.

\begin{figure}[H]
  \centering
  \begin{subfigure}[b]{0.48\linewidth}
    \centering
    \includegraphics[width=\linewidth]{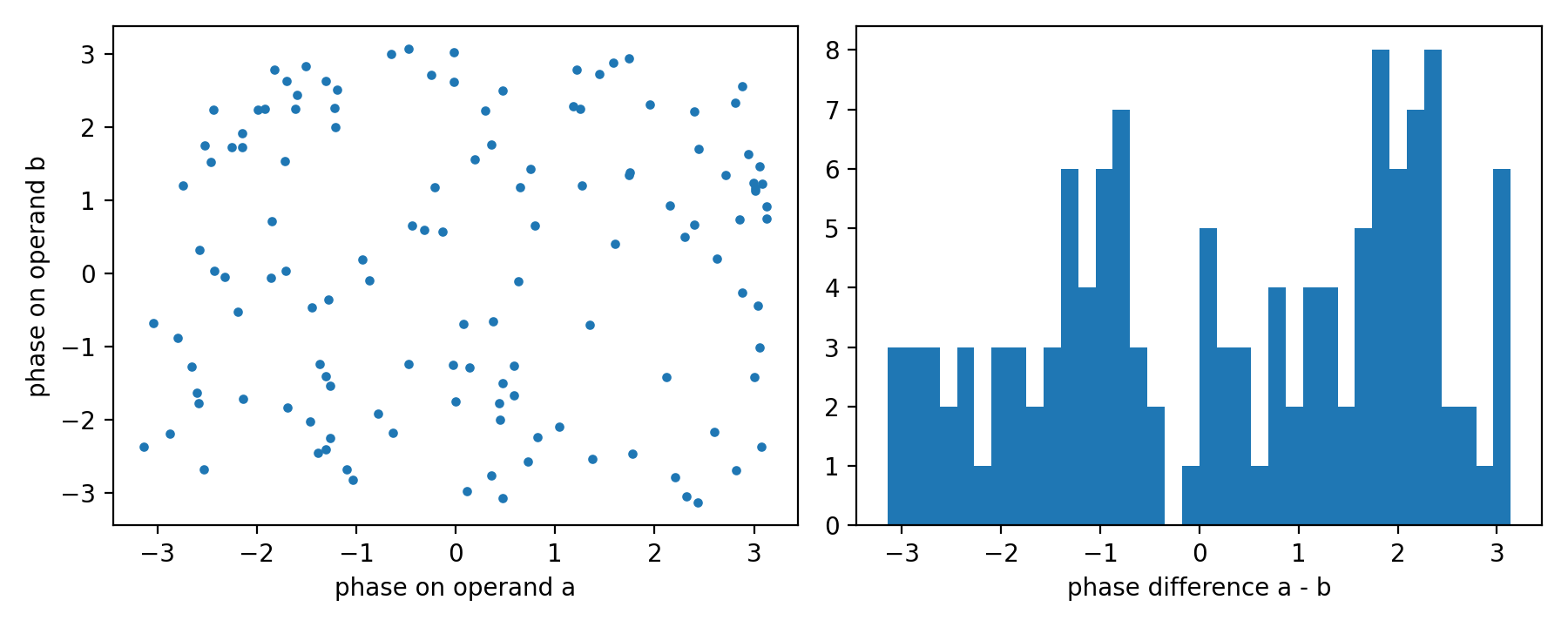}
    \caption{Phase alignment.}
    \label{fig:phase}
  \end{subfigure}\hfill
  \begin{subfigure}[b]{0.48\linewidth}
    \centering
    \includegraphics[width=\linewidth]{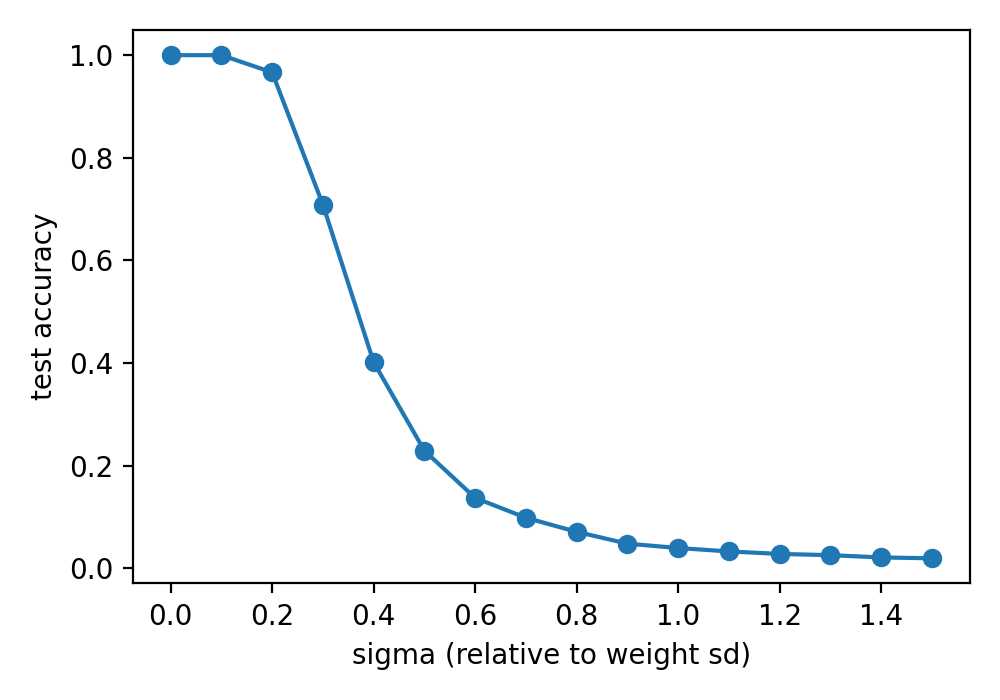}
    \caption{Weight perturbation robustness.}
    \label{fig:noise}
  \end{subfigure}
  \caption{\textbf{Phase structure and robustness of the MLP solution} (seed 0).}
\end{figure}

\paragraph{Phase structure and robustness.}
For each of the 128 hidden units, the phases of its dominant frequency on operands $a$ and $b$ are spread over the whole circle, and their differences are close to uniform (Figure~\ref{fig:phase}). There are no discrete phase families, consistent with the analytical solutions of \citet{gromov2024grokking}, in which each unit's phases are free parameters. Adding Gaussian noise with standard deviation $\sigma$ times the standard deviation of each weight matrix to $W_1$ and $W_2$ lowers test accuracy to 71\% at $\sigma=0.3$, 14\% at $\sigma=0.6$ and 5\% at $\sigma=0.9$ (Figure~\ref{fig:noise}, seed 0, step $10^5$).

\section{Timing, Replication and Rank Interventions}
\label{app:timing_all}

\subsection{Timing of Compression}
\label{app:tf_timing}

Table~\ref{tab:timing} summarizes the timing of compression in both architectures.

\begin{figure}[H]
  \centering
  \includegraphics[width=0.85\linewidth]{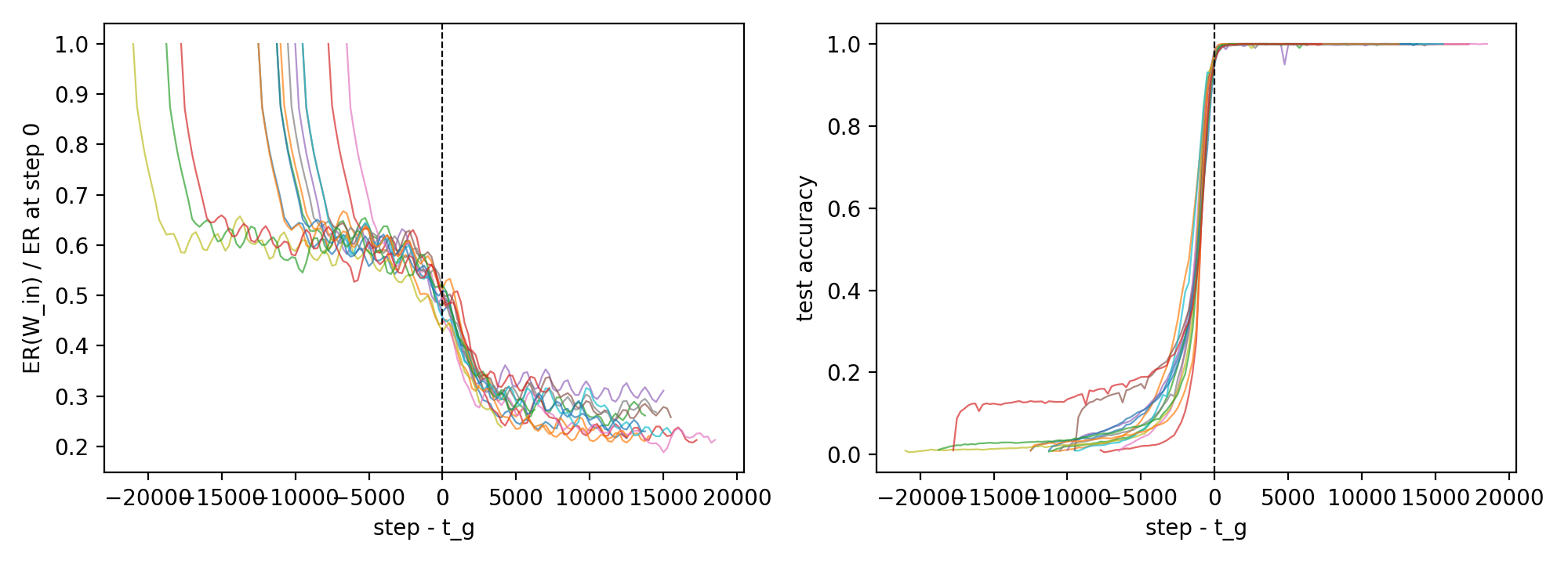}
  \caption{\textbf{Effective rank of $W_\text{in}$} (relative to step 0) \textbf{and test accuracy} of the 14 grokked transformer seeds against steps relative to $t_g$. The effective rank falls steeply while the random initialization decays, plateaus at about 60\% of its initial value, and drops again around $t_g$.}
  \label{fig:tf_aligned}
\end{figure}

\begin{figure}[H]
  \centering
  \includegraphics[width=0.8\linewidth]{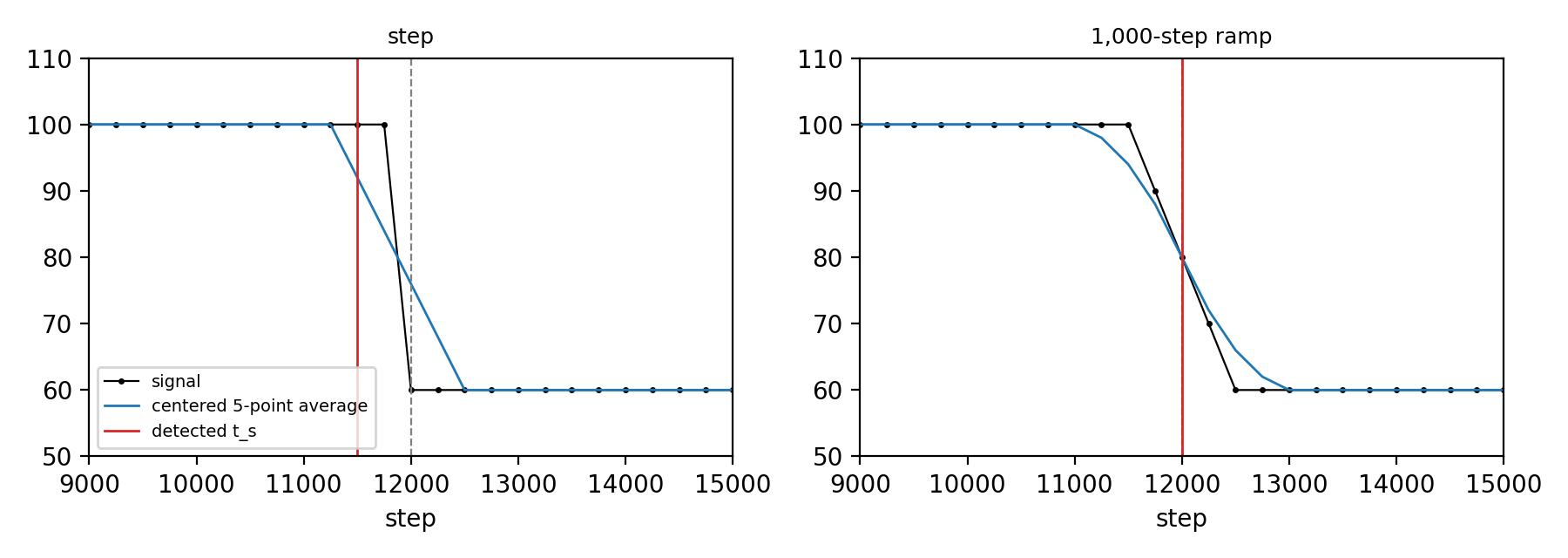}
  \caption{\textbf{Synthetic check of the steepest-drop measure} (logging every 250 steps). Left: a sharp step at step 12{,}000 is located two logging intervals early, because the derivative of the centered average is tied over several steps and the earliest is taken. Right: a 1{,}000-step ramp centered on step 12{,}000 is located exactly.}
  \label{fig:smoothing}
\end{figure}

\paragraph{Steepest-decrease measure.}
The steepest decrease used in Section~\ref{sec:causal} is the step of the most negative central-difference derivative of the effective rank after a centered moving average over five logged steps (\texttt{np.convolve} in \texttt{same} mode, then \texttt{np.gradient}). Each smoothed value belongs to the logged step at its center, so the smoothing adds no delay. The zero-padded average distorts the last two logged steps and the central difference a third, so we exclude the last four. On synthetic signals the measure places a ramp exactly and a sharp step two logging intervals early (Figure~\ref{fig:smoothing}), so any bias shortens the measured lag. It falls after $t_g$ in 13 of 14 transformer seeds of 0--14 (the exception: 250 steps before $t_g$) and in 14 of 15 held-out seeds (the fifteenth at $t_g$). Figure~\ref{fig:tf_aligned} shows the effective rank of $W_\text{in}$ and the test accuracy of the 14 grokked transformer seeds aligned at $t_g$. Counting the decay of the initialization, $31\pm4\%$ of the decrease of $W_\text{in}$ happens after $t_g$ ($W_E$: 22\%, $W_O$: 27\%, $W_U$: 37\%, $W_\text{out}$: 38\%, $W_V$: 62\%).

\begin{table}[tbp]
\centering
\small
\setlength{\tabcolsep}{4pt}
\caption{\textbf{Timing of compression} (mean $\pm$ s.d., in steps). $t_c$ is the first step at which ER falls below 99\% of its running maximum. $t_\text{on}$ and $t_s$ are the start and the steepest point of the effective-rank drop of $W_\text{in}$ after the initialization decay. $t_{10}$ is the first step with test accuracy $\geq10\%$.}
\label{tab:timing}
\begin{tabularx}{\linewidth}{Xcc}
\toprule
 & MLP (15 seeds) & Transformer (29 grokked) \\
\midrule
Memorization step $t_\text{fit}$ & $9{,}133\pm297$ & 250 in all seeds \\
Grokking onset $t_g$ & $12{,}633\pm297$ & $11{,}914\pm4{,}116$ \\
Start of compression $-\,t_g$ & $t_c$: $6{,}367\pm3{,}248$ (15/15 $>0$) & $t_\text{on}$: $-1{,}888\pm992$ (28/29 after $t_{10}$) \\
Steepest ER decrease $-\,t_g$ & $9{,}667\pm3{,}320$ (15/15 $>0$) & $905\pm506$ (27/29 $>0$) \\
Share of ER decrease after $t_g$ & $0.99\pm0.02$ & $0.63\pm0.09^{a}$ \\
\bottomrule
\end{tabularx}
\\[2pt]
{\footnotesize $^{a}$From the plateau after the initialization decay. Including the decay, $0.32\pm0.05$.}
\end{table}

\begin{figure}[H]
  \centering
  \begin{subfigure}[b]{0.3\linewidth}
    \includegraphics[width=\linewidth]{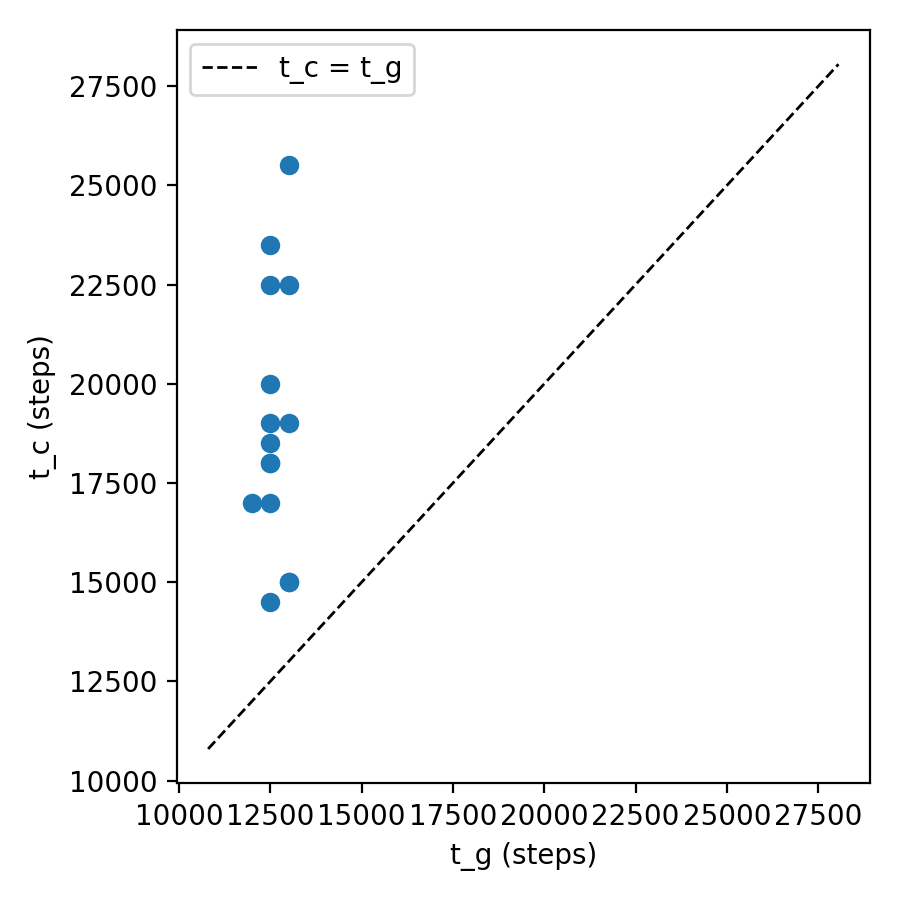}
    \caption{$t_c$ vs.\ $t_g$: all 15 seeds above the diagonal.}
    \label{fig:leadlag_scatter}
  \end{subfigure}\hfill
  \begin{subfigure}[b]{0.65\linewidth}
    \includegraphics[width=\linewidth]{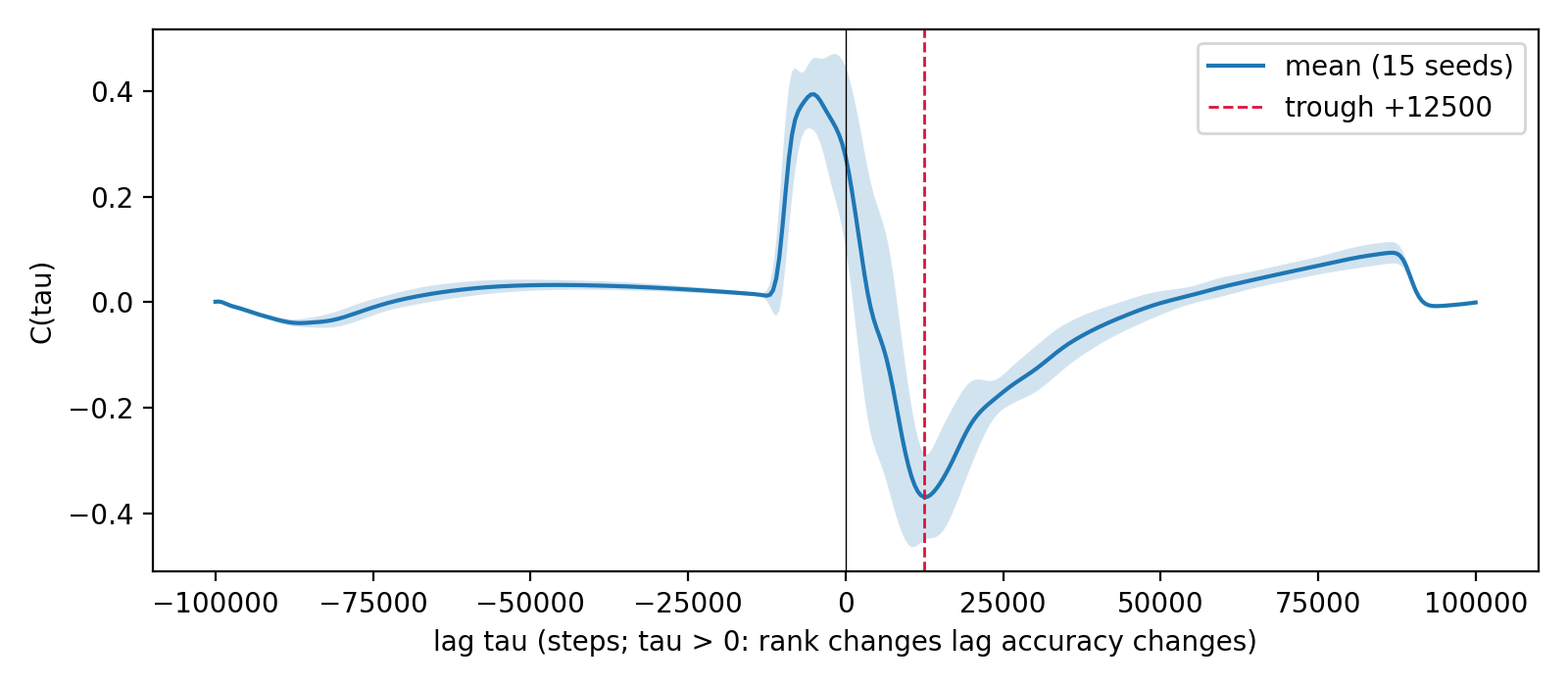}
    \caption{Mean cross-correlation ($\pm1$\,s.d.) of $\partial\mathrm{ER}/\partial t$ and $\partial\mathrm{Acc}/\partial t$.}
    \label{fig:leadlag_xcorr}
  \end{subfigure}
  \caption{\textbf{Compression lags grokking} in every MLP seed.}
  \label{fig:leadlag}
\end{figure}

\paragraph{Lead and lag in the MLP.}
Figure~\ref{fig:leadlag} shows that the compression onset follows the grokking onset in every MLP seed, and that the cross-correlation of the derivatives of effective rank and test accuracy has its trough at $+12{,}500$ steps.

\paragraph{Granger tests.}
Granger causality \citep{granger1969investigating} tests whether the past of one time series improves the prediction of another beyond that series' own past. It measures temporal predictive asymmetry, not interventional causation. We use the first differences of effective rank and test accuracy, logged every 500 steps, with lag order $L\leq4$ chosen by the Akaike information criterion. Augmented Dickey--Fuller tests find these differences stationary in only 8 and 13 of 15 seeds, so we repeat the test after differencing further until the series are stationary in every seed. Changes in test accuracy predict changes in effective rank in every seed in both versions, and the reverse holds in at most 6 of 15 (Table~\ref{tab:granger}). Since test accuracy is flat at 100\% for most of training, these tests largely restate the order of the two pulses.

\begin{table}[H]
\centering
\small
\caption{\textbf{Granger tests} (temporal predictive asymmetry) across 15 MLP seeds.}
\label{tab:granger}
\begin{tabular}{llcccc}
\toprule
Series & Null hypothesis & $\bar{F}$ & $\bar{p}$ & Med.\ $p$ & Rejected ($p<0.05$) \\
\midrule
First differences & Compression $\not\to$ Grokking & $2.84$  & $0.133$ & $0.059$ & 6/15 \\
                  & Grokking $\not\to$ Compression & $24.77$ & $0.003$ & $<0.001$ & 15/15 \\
\midrule
Stationary (ADF) & Compression $\not\to$ Grokking & $2.44$  & $0.257$ & $0.126$ & 4/15 \\
                 & Grokking $\not\to$ Compression & $19.77$ & $0.003$ & $<0.001$ & 15/15 \\
\bottomrule
\end{tabular}
\end{table}

\subsection{Networks That Memorize Without Grokking}
\label{app:fit}

\paragraph{Transformers.}
Table~\ref{tab:neg_timing} compares the second effective-rank drop of $W_\text{in}$ in the 29 grokked transformer seeds with the seven runs that memorize the training set without generalizing. All runs memorize the training set by step 250. In grokked seeds, the effective rank of $W_\text{in}$ falls by $60\pm5\%$ between step 4{,}000 and the end of training, with a peak rate of $10.4\pm2.0$ units per 1{,}000 steps. In non-grokked runs it falls by 15--23\%, at 2.1--4.6 units per 1{,}000 steps.

\begin{table}[H]
\centering
\small
\caption{\textbf{Effective rank of $W_\text{in}$ after the initialization decay} (transformer). Peak rate: steepest smoothed decrease after step 4{,}000.}
\label{tab:neg_timing}
\begin{tabular}{lccc}
\toprule
Run & Final test acc.\ (\%) & Peak rate (per 1{,}000 steps) & Relative change, step 4{,}000 $\to$ end \\
\midrule
Grokked (29 seeds) & $\geq95$ & $-10.4\pm2.0$ & $-0.60\pm0.05$ \\
wd 1.0, seed 12 & 8.2 & $-4.0$ & $-0.15$ \\
wd 0.1, seeds 0--2 & 1.8--4.3 & $-2.1$ to $-2.3$ & $-0.22$ to $-0.23$ \\
wd 0.3, seeds 0--2 & 9.1--13.2 & $-4.3$ to $-4.6$ & $-0.17$ to $-0.19$ \\
\bottomrule
\end{tabular}
\end{table}

\paragraph{Start of the drop.}
The start $t_\text{on}$ of the drop is the first logged step of the contiguous stretch before $t_s$ in which the smoothed decrease is at least 20\% of its peak rate. It falls between the steps at which test accuracy first exceeds 10\% and 50\% in most seeds: $t_\text{on}-t_{10}=3{,}095\pm3{,}778$ steps (28 of 29 non-negative, median 1{,}750) and $t_\text{on}-t_{50}=-828\pm964$ (4 of 29 non-negative). Three seeds have $t_{10}=500$ from an early fluctuation of test accuracy, which inflates the mean of the first difference. Test accuracy at $t_\text{on}$ is $34\pm22\%$, and 37\% of the post-plateau decrease happens before $t_g$.

\paragraph{Matched steps.}
Table~\ref{tab:matched} gives the variance counts at steps 10{,}000--15{,}000. At step 5{,}000, before any seed has generalized, the 29 grokked seeds and the non-grokked seed at the same weight decay have nearly the same counts ($73.5\pm3.9$ against 75 for $W_\text{in}$, $65.2\pm2.1$ against 70 for $W_U$), while the runs at weight decay 0.1 and 0.3 have higher counts (113 and 92 for $W_\text{in}$), a weight-decay effect. At step 20{,}000, 28 seeds have generalized and their counts are $32.3\pm4.7$, $21.4\pm7.0$, $15.2\pm7.0$ and $24.5\pm5.3$ for $W_E$, $W_\text{in}$, $W_\text{out}$ and $W_U$, against 48, 78, 68 and 64 for the non-grokked seed.

\begin{table}[tbp]
\centering
\small
\setlength{\tabcolsep}{3pt}
\caption{\textbf{Low rank is acquired at the transition} (transformer, weight decay 1.0). Singular directions carrying 99\% of $\sum_i\sigma_i^2$ at the same training step, for grokked seeds before and after their own onset and for the seed that never groks. Mean $\pm$ s.d., with the number of seeds in parentheses.}
\label{tab:matched}
\begin{tabular}{rlcccc}
\toprule
Step & Group & $W_E$ & $W_\text{in}$ & $W_\text{out}$ & $W_U$ \\
\midrule
10{,}000 & before own onset (17) & $47.8\pm2.4$ & $77.2\pm8.0$ & $64.1\pm4.5$ & $60.4\pm4.4$ \\
 & after own onset (12) & $39.0\pm4.6$ & $42.8\pm12.3$ & $33.2\pm12.3$ & $34.8\pm7.1$ \\
 & never groks (1) & 52 & 84 & 71 & 67 \\
12{,}500 & before own onset (9) & $44.3\pm2.8$ & $72.2\pm7.4$ & $61.7\pm3.6$ & $60.2\pm3.7$ \\
 & after own onset (20) & $36.0\pm4.1$ & $32.1\pm13.3$ & $24.2\pm15.1$ & $30.4\pm9.5$ \\
 & never groks (1) & 50 & 86 & 70 & 66 \\
15{,}000 & before own onset (8) & $43.5\pm2.6$ & $67.2\pm8.5$ & $59.1\pm3.2$ & $57.2\pm2.7$ \\
 & after own onset (21) & $32.1\pm4.8$ & $22.4\pm3.4$ & $15.0\pm5.1$ & $24.7\pm5.3$ \\
 & never groks (1) & 45 & 81 & 67 & 65 \\
\bottomrule
\end{tabular}
\end{table}

\begin{figure}[H]
  \centering
  \includegraphics[width=0.85\linewidth]{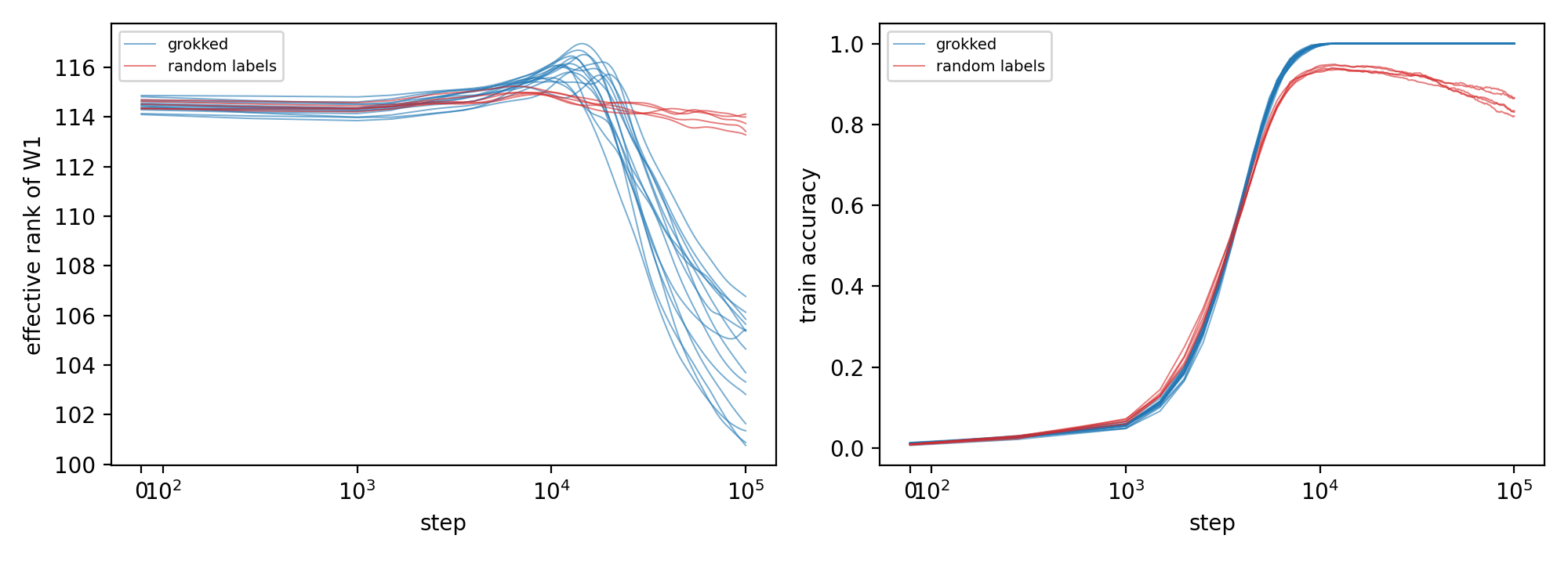}
  \caption{\textbf{MLPs that memorize without generalizing barely compress.} Effective rank of $W_1$ (left) and training accuracy (right) for 15 grokked MLPs (blue) and 5 random-label MLPs (red).}
  \label{fig:mlp_random}
\end{figure}

\paragraph{MLPs trained on random labels.}
Figure~\ref{fig:mlp_random} shows the effective rank and training accuracy of the five random-label MLPs next to the grokked ones. The random-label networks reach 90\% training accuracy at step $7{,}300\pm274$, end at $84\pm2\%$, and their effective rank falls from $115.0\pm0.1$ to $113.7\pm0.4$ ($1.1\pm0.3\%$), against $116.1\pm0.4$ to $104.0\pm2.1$ ($10.5\pm1.8\%$) in grokked networks.

\subsection{Replication on Held-Out Seeds and Tasks}
\label{app:prereg}

\paragraph{Data.}
H1 uses the transformer exactly as in the main experiments (weight decay 1.0, 25,000 steps) with seeds 15--29, which were trained only after the timing measure had been defined on seeds 0--14. H2--H5 use the MLP exactly as in the main runs, on $ab\bmod97$ and $(a{+}b)\bmod113$, with seeds 0--14 per task. H6 uses the same MLP with $W_1=AB$, $r\in\{128,64,32\}$, seeds 0--2 per task and rank. The analysis scripts are in the code repository.

\paragraph{Definitions.}
$t_g$ is the first logged step with test accuracy $\geq95\%$. A run counts as grokked if its final test accuracy is $\geq95\%$, and only grokked runs enter the tests. $t_s$ is the logged step with the most negative derivative of the five-point moving average of the effective rank, over logged steps from step 4,000 (transformer, $W_\text{in}$) or 1,000 (MLP, $W_1$) up to four logging intervals before the end. $t_c$ is the first logged step $\geq1{,}000$ at which the effective rank of $W_1$ falls below 99\% of its running maximum. $d_{0.99}(s)$ is the smallest $k$ whose top-$k$ truncation of $W_1$ keeps $\geq99\%$ of the test accuracy of checkpoint $s$. $s_g$ is the first saved checkpoint $\geq t_g$, $s_+$ the first saved checkpoint $\geq t_g+5{,}000$ that is not the final one, and $s_\text{end}$ the final checkpoint.

\paragraph{Tests.}
All tests are one-sided at $\alpha=0.05$ and run separately per task, and every result is reported. H1: $t_s-t_g>0$ (one-sample $t$-test, with the number of seeds with $t_s>t_g$ and an exact sign test). H2: $t_c-t_g>0$ and H3: $t_s-t_g>0$ (one-sample $t$-tests). H4: $d_{0.99}(s_g)>d_{0.99}(s_+)$ and H5: $d_{0.99}(s_\text{end})>d_{0.99}(s_+)$ (paired $t$-tests). H6 (descriptive): the factorized full-rank model groks earlier than the unfactorized one, and $t_g$ increases as $r$ decreases from 128 to 64 to 32.

\paragraph{Results.}
Every test is significant at $\alpha=0.05$ (Tables~\ref{tab:prereg}, \ref{tab:prereg_h1} and~\ref{tab:prereg_tasks}, and Table~\ref{tab:prereg_rank} for H6). With a Bonferroni correction over the nine $t$-tests, H4 on $(a{+}b)\bmod113$ is not. For H1, the 95\% confidence interval of $t_s-t_g$ is $[560, 1{,}140]$ steps (median 750), and the exact sign test gives $p=4.9\times10^{-4}$. Twice the number of frequencies holding 99\% of the Fourier power of $W_1$ is $90.0\pm2.4$ for $ab\bmod97$ (Fourier transform over the nonzero residues ordered by the discrete logarithm base 5) against $d_{0.99}=89.8\pm2.0$, but $105.5\pm2.1$ for $(a{+}b)\bmod113$ against $d_{0.99}=85.5\pm3.2$, so the frequency count does not predict $d_{0.99}$ in general (\S\ref{sec:comparison}). Multiplication loses a little test accuracy late in training, from 100.0\% at step 30,000 to $99.7\pm0.3\%$ at the end. $d_\text{out}$, which H4 and H5 do not use, does not narrow on either task: it is 121.7--127.4 at every checkpoint (Table~\ref{tab:baselines}).

\begin{table}[H]
\centering
\footnotesize
\setlength{\tabcolsep}{3pt}
\caption{\textbf{Replication tests} (one-sided tests, mean $\pm$ s.d.\ in steps or directions, with the number of seeds showing a positive difference in parentheses).}
\label{tab:prereg}
\begin{tabularx}{\linewidth}{lXccc}
\toprule
 & Quantity & Transformer, held-out & MLP, $ab\bmod97$ & MLP, $(a{+}b)\bmod113$ \\
\midrule
H1 & $t_s-t_g$ ($W_\text{in}$) & $850\pm524$ (14/15)$^{*}$ & -- & -- \\
H2 & $t_c-t_g$ & -- & $6{,}300\pm1{,}461$ (15/15)$^{*}$ & $6{,}700\pm4{,}519$ (15/15)$^{*}$ \\
H3 & $t_s-t_g$ ($W_1$) & -- & $9{,}800\pm2{,}274$ (15/15)$^{*}$ & $11{,}867\pm6{,}741$ (15/15)$^{*}$ \\
H4 & $d_{0.99}(s_g)-d_{0.99}(s_+)$ & -- & $32.9\pm13.7$ (15/15)$^{*}$ & $10.8\pm16.8$ (12/15)$^{*}$ \\
H5 & $d_{0.99}(s_\text{end})-d_{0.99}(s_+)$ & -- & $25.1\pm2.9$ (15/15)$^{*}$ & $16.9\pm3.1$ (15/15)$^{*}$ \\
H6 & $t_g$ for $AB$ at $r=128/64/32$, and unfactorized & -- & $2{,}000/2{,}667/4{,}667$ and $12{,}833$ & $2{,}000/3{,}000/4{,}500$ and $15{,}333$ \\
\bottomrule
\end{tabularx}
\\[2pt]
{\footnotesize $^{*}p<0.05$. The largest $p$ is $0.013$ (H4, $(a{+}b)\bmod113$). H1--H3 use one-sample $t$-tests, H4 and H5 paired $t$-tests, and H6 is descriptive, with 3 seeds per arm. $s_g$ is the first checkpoint at or after $t_g$, $s_+$ the first checkpoint at least 5,000 steps later, and $s_\text{end}$ the final checkpoint.}
\end{table}

\begin{table}[H]
\centering
\footnotesize
\setlength{\tabcolsep}{6pt}
\caption{\textbf{H2--H5 on the two further MLP tasks} (15 seeds each, mean $\pm$ s.d.). Below each test, the number of seeds with a positive difference, the $t$ statistic and the one-sided $p$.}
\label{tab:prereg_tasks}
\begin{tabular}{lcc}
\toprule
 & $ab\bmod97$ & $(a{+}b)\bmod113$ \\
\midrule
Grokked & 15/15 & 15/15 \\
$t_g$ & $12{,}633\pm297$ & $15{,}567\pm417$ \\
$t_c$ & $18{,}933\pm1{,}498$ & $22{,}267\pm4{,}629$ \\
\midrule
H2: $t_c-t_g$ & $6{,}300\pm1{,}461$ & $6{,}700\pm4{,}519$ \\
 & {\scriptsize 15/15, $t=16.70$, $p=6.1\times10^{-11}$} & {\scriptsize 15/15, $t=5.74$, $p=2.6\times10^{-5}$} \\
\addlinespace
H3: $t_s-t_g$ & $9{,}800\pm2{,}274$ & $11{,}867\pm6{,}741$ \\
 & {\scriptsize 15/15, $t=16.69$, $p=6.2\times10^{-11}$} & {\scriptsize 15/15, $t=6.82$, $p=4.2\times10^{-6}$} \\
\addlinespace
H4: $d_{0.99}(s_g)-d_{0.99}(s_+)$ & $32.9\pm13.7$ & $10.8\pm16.8$ \\
 & {\scriptsize 15/15, $t=9.31$, $p=1.1\times10^{-7}$} & {\scriptsize 12/15, $t=2.49$, $p=0.013$} \\
\addlinespace
H5: $d_{0.99}(s_\text{end})-d_{0.99}(s_+)$ & $25.1\pm2.9$ & $16.9\pm3.1$ \\
 & {\scriptsize 15/15, $t=33.01$, $p=5.6\times10^{-15}$} & {\scriptsize 15/15, $t=20.98$, $p=2.8\times10^{-12}$} \\
\midrule
Share of the ER$(W_1)$ decrease after $t_g$ & $1.00\pm0.01$ & $0.98\pm0.02$ \\
$d_{0.99}$ at the end & $89.8\pm2.0$ & $85.5\pm3.2$ \\
Top / bottom / random-subspace truncation (\%) & 99.10 / 36.84 / 20.68 & 99.08 / 27.93 / 33.03 \\
\bottomrule
\end{tabular}
\\[2pt]
{\footnotesize Truncations keep $k=d_{0.99}$ directions. Chance accuracy: 1.03\% ($p=97$) and 0.88\% ($p=113$). For $(a{+}b)\bmod113$, $s_g$ is step 20,000 in 12 seeds. The three seeds with $s_g$ at step 15,000 narrow from 106, 111 and 115 directions to 66, 71 and 67.}
\end{table}

\begin{table}[H]
\centering
\small
\caption{\textbf{H1 per seed:} held-out transformer seeds, $t_g$ and $t_s-t_g$ (steps).}
\label{tab:prereg_h1}
\begin{tabular}{rrr@{\hspace{18pt}}rrr@{\hspace{18pt}}rrr}
\toprule
Seed & $t_g$ & $t_s-t_g$ & Seed & $t_g$ & $t_s-t_g$ & Seed & $t_g$ & $t_s-t_g$ \\
\midrule
15 & 7,000  & $+500$   & 20 & 17,500 & $+500$   & 25 & 17,750 & $+250$   \\
16 & 12,500 & $+1{,}000$ & 21 & 8,000  & $0$      & 26 & 8,750  & $+500$   \\
17 & 7,500  & $+750$   & 22 & 13,000 & $+1{,}000$ & 27 & 15,750 & $+750$   \\
18 & 17,000 & $+1{,}000$ & 23 & 16,500 & $+1{,}750$ & 28 & 8,500  & $+750$   \\
19 & 7,500  & $+750$   & 24 & 10,250 & $+2{,}000$ & 29 & 8,250  & $+1{,}250$ \\
\bottomrule
\end{tabular}
\end{table}

\subsection{Forcing Low Rank}
\label{app:rank}

\begin{figure}[H]
  \centering
  \begin{subfigure}[b]{0.49\linewidth}
    \includegraphics[width=\linewidth]{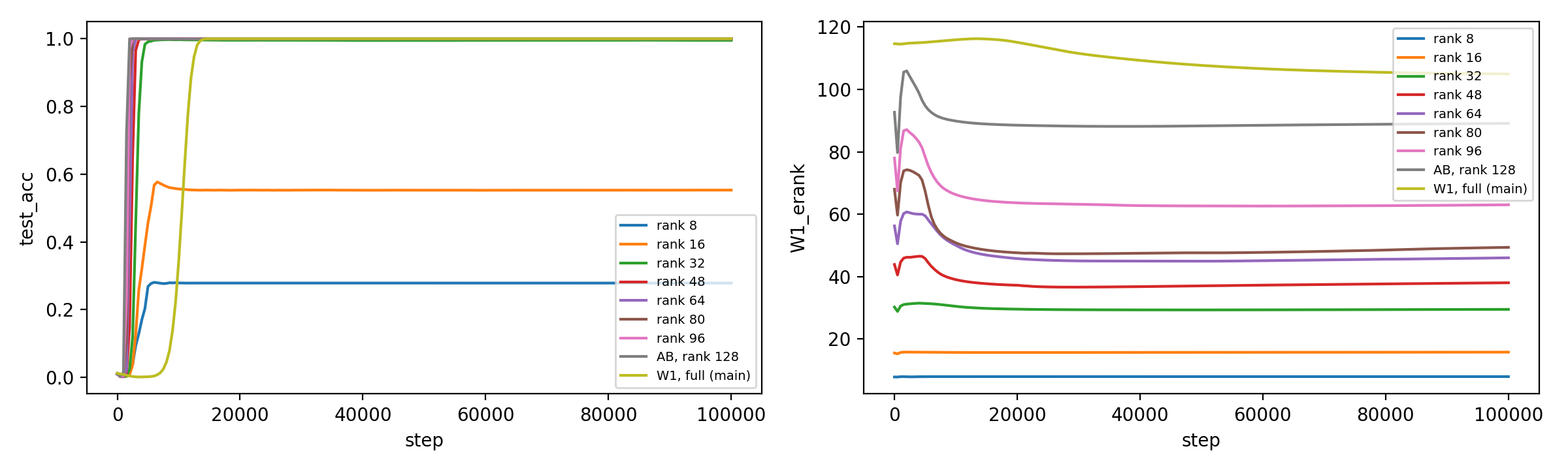}
    \caption{Rank-constrained parameterization $W_1=AB$.}
  \end{subfigure}\hfill
  \begin{subfigure}[b]{0.49\linewidth}
    \includegraphics[width=\linewidth]{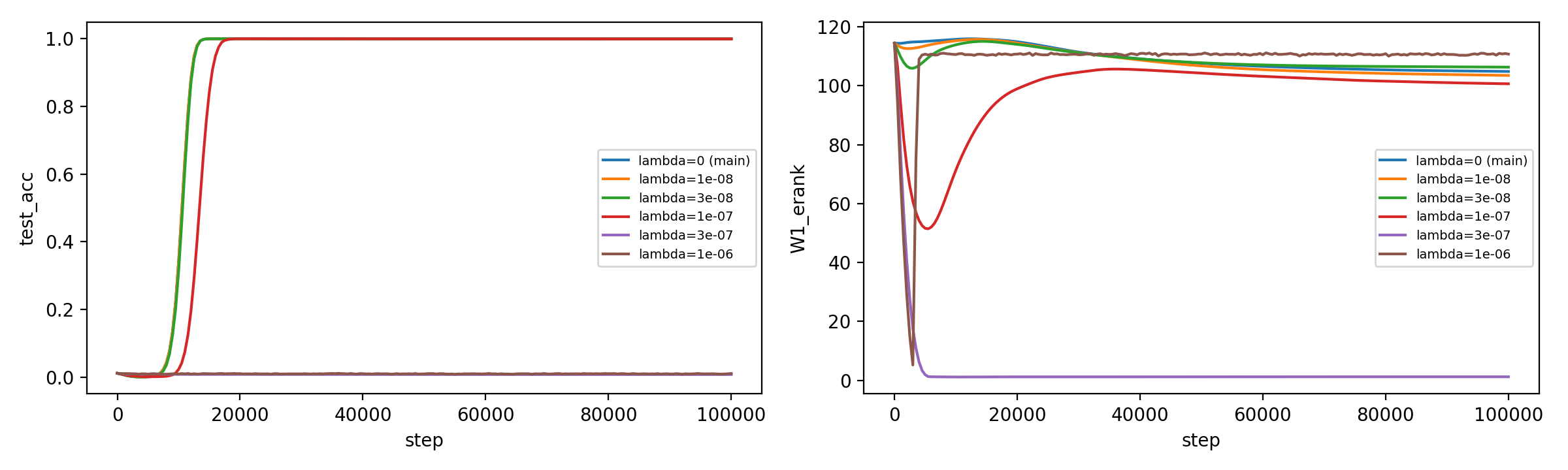}
    \caption{Nuclear-norm penalty from step 0.}
  \end{subfigure}
  \caption{\textbf{Forcing low rank from initialization} (mean test accuracy and effective rank of $W_1$ against training step).}
  \label{fig:rank_interventions}
\end{figure}

\begin{figure}[H]
  \centering
  \includegraphics[width=0.6\linewidth]{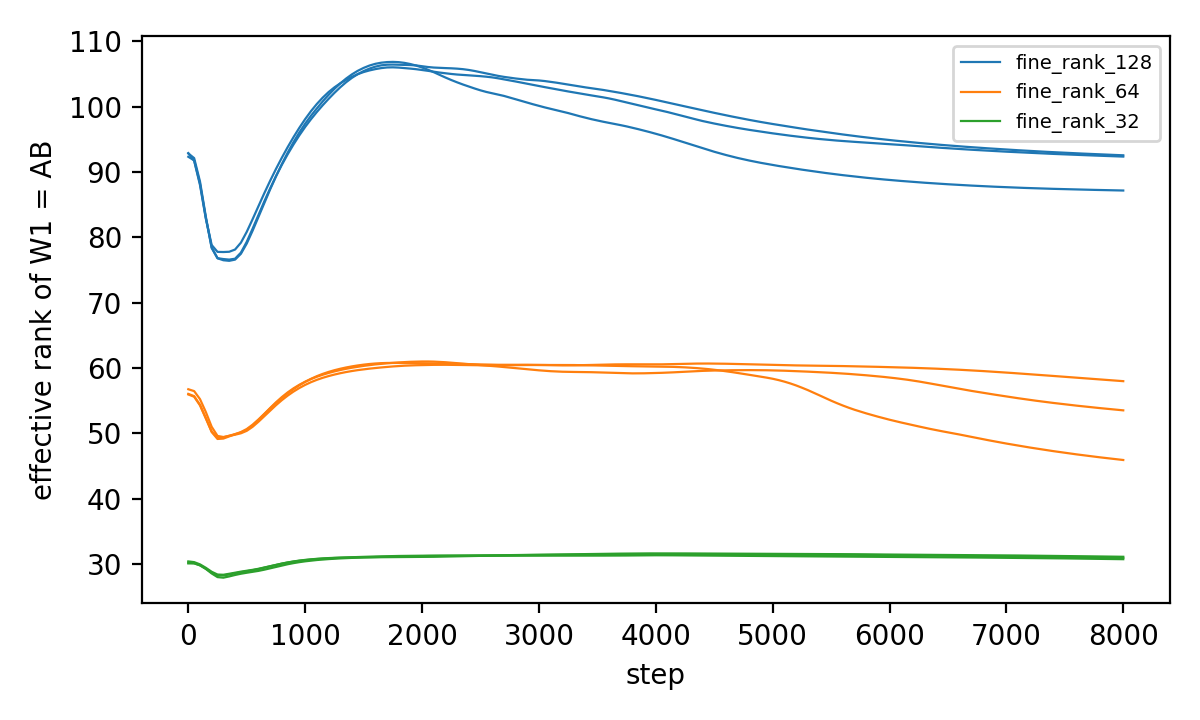}
  \caption{\textbf{Effective rank of factorized MLPs} ($W_1=AB$ with $r=128$, 64 and 32, logged every 50 steps, 3 seeds each).}
  \label{fig:fine}
\end{figure}

Table~\ref{tab:bottleneck} and Figure~\ref{fig:rank_interventions} report the MLP rank interventions summarized in Section~\ref{sec:bottleneck}. Table~\ref{tab:factor_er} shows that the factorization alone biases $W_1$ toward low rank, and Table~\ref{tab:prereg_rank} repeats the factorization on the two further tasks. Table~\ref{tab:fine} and Figure~\ref{fig:fine} repeat the factorized control and two rank caps with logging every 50 steps, on the same seeds as above. The full-rank factorization groks $7.5\pm0.1$, $7.2\pm0.1$ and $8.4\pm0.2$ times sooner than the unfactorized model on the three tasks. Its effective rank dips early, rises to its maximum at the onset and falls afterwards.

\begin{table}[H]
\centering
\small
\caption{\textbf{Forcing low rank from initialization} (MLP). Rank-constrained models use 3 seeds and the penalty 5 seeds.}
\label{tab:bottleneck}
\begin{tabular}{lcccc}
\toprule
Model & Grokked & $t_g$ & Final test acc.\ (\%) & Final ER$(W_1)$ \\
\midrule
Unfactorized $W_1$ (main)    & 3/3 & $12{,}833\pm289$ & $100.0\pm0.0$ & $104.9\pm1.1$ \\
$W_1=AB$, $r=128$ (control)  & 3/3 & $2{,}000\pm0$    & $99.9\pm0.2$  & $89.1\pm0.9$ \\
$W_1=AB$, $r=96$             & 3/3 & $2{,}167\pm289$  & $100.0\pm0.0$ & $63.0\pm5.7$ \\
$W_1=AB$, $r=80$             & 3/3 & $2{,}500\pm0$    & $100.0\pm0.0$ & $49.4\pm1.2$ \\
$W_1=AB$, $r=64$             & 3/3 & $2{,}667\pm289$  & $100.0\pm0.0$ & $46.0\pm3.7$ \\
$W_1=AB$, $r=48$             & 3/3 & $3{,}167\pm289$  & $100.0\pm0.0$ & $38.0\pm0.6$ \\
$W_1=AB$, $r=32$             & 3/3 & $4{,}500\pm0$    & $99.5\pm0.3$  & $29.5\pm0.1$ \\
$W_1=AB$, $r=16$             & 0/3 & --               & $55.3\pm10.6$ & $15.8\pm0.0$ \\
$W_1=AB$, $r=8$              & 0/3 & --               & $27.8\pm6.1$  & $8.0\pm0.0$ \\
\midrule
No penalty                   & 5/5 & $12{,}700\pm447$ & $100.0\pm0.0$ & $104.9\pm1.6$ \\
$\lambda=10^{-8}$            & 5/5 & $12{,}800\pm274$ & $100.0\pm0.0$ & $103.5\pm2.2$ \\
$\lambda=3\times10^{-8}$     & 5/5 & $12{,}800\pm274$ & $100.0\pm0.1$ & $106.3\pm1.3$ \\
$\lambda=10^{-7}$            & 5/5 & $16{,}200\pm447$ & $100.0\pm0.0$ & $100.7\pm3.4$ \\
$\lambda=3\times10^{-7}$     & 0/5 & --               & $0.8\pm0.1$   & $1.2\pm0.3$ \\
$\lambda=10^{-6}$            & 0/5 & --               & $1.1\pm0.1$   & $110.8\pm0.4^{\S}$ \\
\bottomrule
\end{tabular}
\\[2pt]
{\footnotesize $^{\S}$At $\lambda=10^{-6}$ the penalty prevents learning. The high effective rank presumably reflects weights shrunk to small, unstructured values, which we did not inspect.}
\end{table}

\begin{table}[H]
\centering
\small
\caption{\textbf{The factorization biases $W_1$ toward low rank.} Effective rank of $W_1=AB$ with $r=128$ and of the unfactorized $W_1$ (seeds 0--2, $(a{+}b)\bmod97$). The factorized model groks at step 2,000, the unfactorized one at $12{,}833\pm289$.}
\label{tab:factor_er}
\begin{tabular}{rcc}
\toprule
Step & $W_1=AB$, $r=128$ & Unfactorized $W_1$ \\
\midrule
0      & $92.7\pm0.3$  & $114.6\pm0.1$ \\
500    & $79.7\pm1.0$  & $114.5\pm0.1$ \\
1,000  & $97.4\pm0.6$  & $114.5\pm0.1$ \\
1,500  & $105.5\pm0.3$ & $114.6\pm0.2$ \\
2,000  & $105.9\pm0.3$ & $114.7\pm0.2$ \\
3,000  & $102.4\pm2.0$ & $114.9\pm0.2$ \\
5,000  & $94.8\pm3.3$  & $115.1\pm0.1$ \\
10,000 & $89.8\pm2.7$  & $115.9\pm0.1$ \\
20,000 & $88.5\pm2.1$  & $115.1\pm0.4$ \\
\bottomrule
\end{tabular}
\end{table}

\begin{table}[H]
\centering
\small
\caption{\textbf{Factorization runs on the two further tasks (H6)} (3 seeds per arm, unfactorized: main seeds 0--2).}
\label{tab:prereg_rank}
\begin{tabular}{llcccc}
\toprule
Task & Model & Grokked & $t_g$ & Final test acc.\ (\%) & Final ER$(W_1)$ \\
\midrule
$ab\bmod97$ & Unfactorized $W_1$ & 3/3 & $12{,}833\pm289$ & $99.7\pm0.4$ & $102.3\pm0.7$ \\
 & $W_1=AB$, $r=128$ & 3/3 & $2{,}000\pm0$ & $99.3\pm0.7$ & $85.6\pm1.4$ \\
 & $W_1=AB$, $r=64$ & 3/3 & $2{,}667\pm289$ & $100.0\pm0.0$ & $47.9\pm0.2$ \\
 & $W_1=AB$, $r=32$ & 3/3 & $4{,}667\pm289$ & $99.0\pm0.5$ & $30.1\pm0.2$ \\
\midrule
$(a{+}b)\bmod113$ & Unfactorized $W_1$ & 3/3 & $15{,}333\pm289$ & $100.0\pm0.0$ & $109.5\pm1.8$ \\
 & $W_1=AB$, $r=128$ & 3/3 & $2{,}000\pm0$ & $100.0\pm0.0$ & $89.9\pm2.1$ \\
 & $W_1=AB$, $r=64$ & 3/3 & $3{,}000\pm0$ & $100.0\pm0.0$ & $47.0\pm5.1$ \\
 & $W_1=AB$, $r=32$ & 3/3 & $4{,}500\pm0$ & $99.2\pm0.9$ & $29.6\pm0.5$ \\
\bottomrule
\end{tabular}
\end{table}

\begin{table}[H]
\centering
\small
\caption{\textbf{Factorized MLPs logged every 50 steps} (3 seeds each, mean $\pm$ s.d.). ER: effective rank of $W_1=AB$.}
\label{tab:fine}
\begin{tabular}{llcccc}
\toprule
Task & Model & $t_g$ & ER minimum before $t_g$ (step) & ER at $t_g$ & ER at step 8{,}000 \\
\midrule
$(a{+}b)\bmod97$ & unfactorized & $12{,}483\pm340$ & -- & $116.2\pm0.2$ & -- \\
 & $r=128$ & $1{,}667\pm58$ & $76.9\pm0.7$ ($333$) & $106.3\pm0.5$ & $90.7\pm3.1$ \\
 & $r=64$ & $2{,}450\pm100$ & $49.3\pm0.2$ ($267$) & $60.5\pm0.1$ & $52.5\pm6.1$ \\
 & $r=32$ & $4{,}133\pm29$ & $28.2\pm0.2$ ($300$) & $31.5\pm0.1$ & $31.0\pm0.2$ \\
$ab\bmod97$ & unfactorized & $12{,}567\pm189$ & -- & $115.6\pm0.6$ & -- \\
 & $r=128$ & $1{,}750\pm0$ & $73.8\pm0.6$ ($300$) & $106.0\pm1.3$ & $87.3\pm1.5$ \\
$(a{+}b)\bmod113$ & unfactorized & $15{,}233\pm293$ & -- & $117.6\pm0.2$ & -- \\
 & $r=128$ & $1{,}817\pm29$ & $78.6\pm0.0$ ($333$) & $108.5\pm0.5$ & $93.4\pm2.3$ \\
\bottomrule
\end{tabular}
\end{table}

\subsection{Label-Free Onset Detection}
\label{app:ipr}

\begin{figure}[H]
  \centering
  \includegraphics[width=0.55\linewidth]{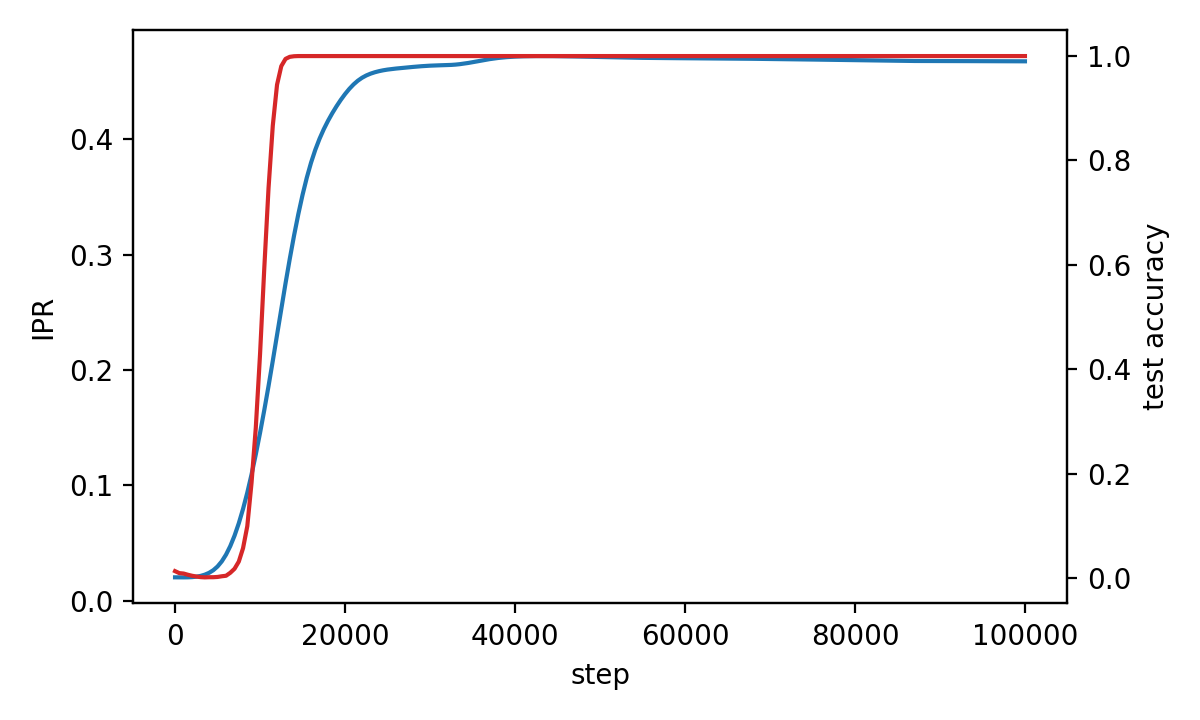}
  \caption{\textbf{Fourier IPR of $W_1$} (blue) and test accuracy (red), MLP seed 0.}
  \label{fig:ipr}
\end{figure}

Table~\ref{tab:ipr_sens} reports the onset detectors summarized in Section~\ref{sec:ipr}, and Figure~\ref{fig:ipr} the Fourier IPR of one MLP seed. In the MLP, the Fourier IPR fires at or just after the end of the baseline window, so its lead is set by the window, $t_g-W$. Spectral IPR, effective rank and stable rank of $W_1$ change after $t_g$ (early in at most 2 of 15 seeds, with median leads from $-8{,}000$ to $-23{,}500$ steps). For the transformer, ``early'' is counted out of the seeds with $t_g>W$, the only ones that can be early. Of 63 metric, window and threshold combinations tested on the MLP, only the gradient norm (window 5,000, threshold $3\sigma$) detected grokking early in all seeds without false alarms. Given the number of combinations, we regard this as exploratory.

\begin{table}[H]
\centering
\small
\caption{\textbf{Label-free onset detection} (threshold $3\sigma$ above the mean of the baseline window $[0,W)$). \emph{At window end}: fires at the first logged step after the window. \emph{False alarms}: fires in runs that never generalize (5 random-label MLP runs and 7 non-grokked transformer runs). Lead $=t_g-t_\text{onset}$ (median, steps).}
\label{tab:ipr_sens}
\begin{tabular}{llrcccr}
\toprule
Model & Metric & $W$ & Early & At window end & False alarms & Lead \\
\midrule
MLP & Fourier IPR ($W_1$) & 2,500  & 15/15 & 14/15 & 5/5 & 10,000 \\
    &                     & 5,000  & 15/15 & 15/15 & 5/5 & 7,500 \\
    &                     & 10,000 & 15/15 & 14/15 & 5/5 & 2,500 \\
MLP & Gradient norm       & 2,500  & 15/15 & 0/15  & 5/5 & 9,000 \\
    &                     & 5,000  & 15/15 & 0/15  & 0/5 & 2,500 \\
    &                     & 10,000 & 3/15  & 0/15  & 3/5 & 0 \\
\midrule
Transformer & Fourier IPR ($W_E$) & 2,500  & 14/14 & 0/14  & 7/7 & 7,625 \\
            &                     & 5,000  & 14/14 & 0/14  & 7/7 & 5,250 \\
            &                     & 10,000 & 9/9   & 6/14  & 7/7 & 1,000 \\
Transformer & Spectral IPR ($W_\text{in}$) & 2,500  & 14/14 & 0/14 & 7/7 & 3,125 \\
            &                     & 5,000  & 14/14 & 0/14  & 7/7 & 1,625 \\
            &                     & 10,000 & 9/9   & 5/14  & 3/7 & 750 \\
\bottomrule
\end{tabular}
\end{table}

\begin{figure}[H]
  \centering
  \includegraphics[width=\linewidth]{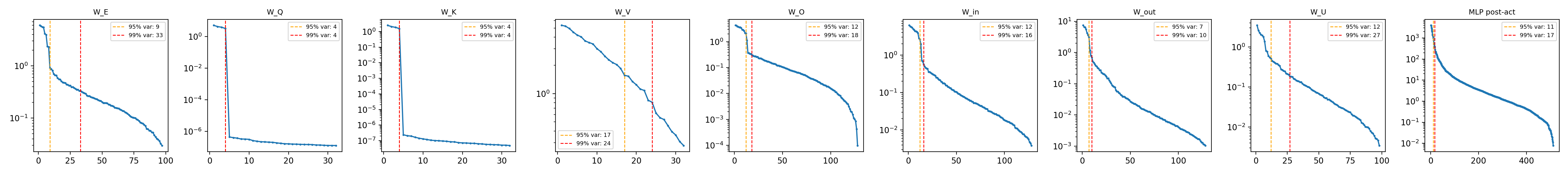}
  \caption{\textbf{Transformer singular value spectra} (seed 0) for $W_E$, $W_Q$, $W_K$, $W_V$, $W_O$, $W_\text{in}$, $W_\text{out}$, $W_U$ and the MLP post-activations, with the 95\% and 99\% variance thresholds.}
  \label{fig:transformer_spectra}
\end{figure}

\section{Transformer: Additional Results}
\label{app:transformer}

\subsection{Dimensionality at the End of Training}

Table~\ref{tab:tf_dims} gives the variance counts and $d_{0.99}$ of every analyzed matrix, and Figure~\ref{fig:transformer_spectra} the singular value spectra for one seed. Averaged over seeds at fixed steps, $d_{0.99}$ mixes seeds before and after their onsets: $d_{0.99}(W_\text{in})$ falls from $18.2\pm7.2$ at step 10,000 (mean test accuracy $57\pm40\%$) to $12.4\pm1.5$ at step 20,000, and $d_{0.99}(W_E)$ from $19.4\pm9.9$ to $7.7\pm1.5$. Aligned at each seed's own onset (Table~\ref{tab:tf_dense}), $d_\text{out}$ narrows during the rise in test accuracy. Table~\ref{tab:comparison} compares the final counts with those of the MLP.

\begin{table}[H]
\centering
\small
\caption{\textbf{Transformer dimensionality} (14 grokked seeds, final checkpoint). $k_{95}$, $k_{99}$: singular directions carrying 95\% and 99\% of $\sum_i\sigma_i^2$. $d_{0.99}$: fewest directions that keep 99\% of test accuracy. Attention projections are stacked across the four heads. Activations are truncated after mean-centering.}
\label{tab:tf_dims}
\begin{tabular}{lcccc}
\toprule
Matrix & Max rank & $k_{95}$ & $k_{99}$ & $d_{0.99}$ \\
\midrule
$W_E$ (embedding)          & 97  & $8.8\pm1.5$  & $30.4\pm4.0$ & $7.7\pm1.5$ \\
$W_Q$ / $W_K$              & 32  & $4.0\pm0.0$  & $4.0\pm0.0$  & $2.6\pm0.5$ / $2.4\pm0.7$ \\
$W_V$                      & 32  & $18.5\pm2.6$ & $24.3\pm2.6$ & $10.9\pm1.5$ \\
$W_O$                      & 128 & $13.2\pm1.4$ & $21.3\pm4.7$ & $12.8\pm1.3$ \\
$W_\text{in}$              & 128 & $13.1\pm1.4$ & $18.1\pm3.8$ & $13.0\pm1.3$ \\
$W_\text{out}$             & 128 & $8.6\pm1.4$  & $11.8\pm2.5$ & $8.7\pm1.5$ \\
$W_U$ (unembedding)        & 98  & $10.3\pm1.3$ & $23.8\pm2.8$ & $8.8\pm1.4$ \\
MLP post-activations       & 512 & $14.4\pm3.3$ & $23.6\pm7.2$ & $17.2\pm3.3$ \\
\bottomrule
\end{tabular}
\end{table}

\begin{table}[H]
\centering
\small
\caption{\textbf{Architecture comparison} (final checkpoints, with 15 MLP seeds, 5 cross-entropy MLP seeds per learning rate and 14 grokked transformer seeds).}
\label{tab:comparison}
\begin{tabularx}{\linewidth}{Xccc}
\toprule
Property & MLP (MSE) & MLP (CE) & Transformer \\
\midrule
Loss, weight decay, training steps & MSE, $10^{-2}$, $10^5$ & CE, $10^{-2}$, $10^5$ & CE, $1.0$, $2.5\times10^4$ \\
$d_{0.99}$ of first-layer weights (of 128) & $89.1\pm2.3$ & $48$--$51$ & $13.0\pm1.3$ \\
$d_\text{out}$ of first-layer weights (of 128) & $127.3\pm0.6$ & $70$--$81$ & $13.0\pm1.9$ \\
When $d_\text{out}$ falls & never & late, after $t_g$ & during the transition \\
Frequencies holding 90\% of first-layer power & $33.7\pm1.9$ ($W_1$) & -- & $4.2\pm0.7$ ($W_E$) \\
\bottomrule
\end{tabularx}
\end{table}

\subsection{Dense Time Course}
\label{app:dense}

All transformer runs were repeated with a checkpoint every 250 steps. For each grokked seed we compute $d_\text{out}$ of $W_E$, $W_\text{in}$, $W_\text{out}$ and $W_U$ at every checkpoint within 3{,}000 steps of its $t_g$, testing every $k$ from 1 upward. The restricted and excluded losses follow \citet{nanda2023progress} in output space: for each input, we keep (restricted) or remove (excluded) the components of the centered residue logits along $\cos(2\pi wc/p)$ and $\sin(2\pi wc/p)$ for the five key frequencies $w$ of $W_E$, and evaluate the cross-entropy on the training set. For each curve, the midpoint is the first checkpoint at which the count falls below the midpoint between its value at the start of the window and its final value. The restricted loss uses the same rule, and the excluded loss the first checkpoint above the midpoint of its rise. The midpoints relative to $t_g$ are: $W_E$ $+26\pm488$ steps (11 of 29 seeds before $t_g$), $W_\text{in}$ $-862\pm503$ (29 of 29), $W_\text{out}$ $-957\pm328$ (29 of 29), $W_U$ $-741\pm375$ (28 of 29), restricted loss $-1{,}966\pm647$ and excluded loss $-1{,}026\pm646$.

Table~\ref{tab:dense_neg} gives $d_\text{out}$ for the runs that never generalize. They narrow slowly, most at weight decay 0.3, where test accuracy also creeps up to 9--13\%, but at the end of training they remain two to nine times wider than the grokked seeds.

\begin{table}[H]
\centering
\small
\caption{\textbf{Runs that never generalize}: $d_\text{out}$ (centered logits within 10\%, test inputs) at step 5{,}000 and at the end of training, with grokked seeds 3{,}000 steps after their onset for comparison.}
\label{tab:dense_neg}
\begin{tabular}{llccccc}
\toprule
Run & Step & $W_E$ & $W_\text{in}$ & $W_\text{out}$ & $W_U$ & Test acc.\ (\%) \\
\midrule
wd 1.0, seed 12 & 5{,}000 & 31 & 33 & 64 & 70 & 2.8 \\
 & end & 23 & 26 & 49 & 55 & 8.2 \\
wd 0.1, seeds 0--2 & 5{,}000 & 45--47 & 42--43 & 74--77 & 81--84 & 1.0--2.8 \\
 & end & 35--38 & 35--37 & 67--69 & 78--83 & 1.8--4.3 \\
wd 0.3, seeds 0--2 & 5{,}000 & 40--42 & 38--40 & 71--73 & 79--82 & 1.2--2.9 \\
 & end & 29--30 & 30--31 & 48--52 & 53--59 & 9.1--13.2 \\
\midrule
Grokked (29), $t_g+3{,}000$ & -- & $8.6\pm2.0$ & $13.8\pm1.7$ & $9.0\pm1.8$ & $8.9\pm1.7$ & 100.0 \\
\bottomrule
\end{tabular}
\end{table}

\subsection{Runs That Never Grok and Attention Projections}

Table~\ref{tab:negatives} lists the seven transformer runs that did not reach 95\% test accuracy within 25,000 steps. $W_Q$ has rank 4 in all of them, and $W_K$ in those trained with weight decay of at least 0.3. At weight decay 0.1, where the random initialization has not yet decayed, $W_K$ keeps 18--19 directions. The rank-4 structure of $W_Q$ and $W_K$, one direction per head, is architectural rather than a signature of grokking: only the attention pattern of the final position, which always holds the ``$=$'' token, affects the output, so each head's query is a single vector and the gradients of its query and key projections have rank one.

\begin{table}[H]
\centering
\small
\caption{Non-grokked transformer runs: final accuracy (\%) and $k_{99}$ (singular directions carrying 99\% of $\sum_i\sigma_i^2$). Grokked seeds (mean): $W_E$ 30.4, $W_Q$ 4.0, $W_K$ 4.0, $W_V$ 24.3, $W_O$ 21.3, $W_\text{in}$ 18.1, $W_\text{out}$ 11.8, $W_U$ 23.8.}
\label{tab:negatives}
\begin{tabular}{lcccccccccc}
\toprule
Run & Train & Test & $W_E$ & $W_Q$ & $W_K$ & $W_V$ & $W_O$ & $W_\text{in}$ & $W_\text{out}$ & $W_U$ \\
\midrule
wd 1.0, seed 12 & 100.0 & 8.2  & 42 & 4 & 4  & 30 & 42 & 68 & 62 & 62 \\
wd 0.1, seed 0  & 100.0 & 4.3  & 43 & 4 & 19 & 31 & 50 & 82 & 78 & 72 \\
wd 0.1, seed 1  & 100.0 & 1.8  & 43 & 4 & 18 & 31 & 51 & 81 & 78 & 73 \\
wd 0.1, seed 2  & 100.0 & 3.0  & 44 & 4 & 18 & 31 & 52 & 84 & 79 & 73 \\
wd 0.3, seed 0  & 100.0 & 13.2 & 34 & 4 & 4  & 31 & 43 & 74 & 61 & 62 \\
wd 0.3, seed 1  & 100.0 & 10.7 & 36 & 4 & 4  & 31 & 43 & 73 & 61 & 62 \\
wd 0.3, seed 2  & 100.0 & 9.1  & 37 & 4 & 4  & 30 & 43 & 77 & 65 & 65 \\
\bottomrule
\end{tabular}
\end{table}

\subsection{Residual-Stream Alignment}

Consecutive components communicate through shared subspaces of the 128-dimensional residual stream, but this is not specific to grokking (Table~\ref{tab:subspace_analysis}). What attention writes is almost exactly what the MLP reads ($0.923\pm0.035$), and what the MLP writes is what the unembedding reads ($0.885\pm0.078$), whereas the MLP's own read and write subspaces are only weakly aligned ($0.330\pm0.050$). The attention-to-MLP alignment is equally high in the non-grokked runs ($0.839$--$0.947$). For a matrix that reads from the residual stream ($W_\text{in}$, $W_U$), its read subspace is spanned by its top-$k$ left singular vectors. For a matrix that writes to it ($W_O$, $W_\text{out}$), its write subspace is spanned by its top-$k$ right singular vectors. For two subspaces with orthonormal bases $U,V\in\mathbb{R}^{128\times k}$, the principal angles are the arccosines of the singular values of $U^\top V$, and we report the mean of their cosines, with $k=11=2\times5$ key frequencies $+1$. The key-frequency subspace of $W_E$ is spanned by the images under $W_E^\top$ of the cosine and sine of each of its five strongest frequencies and of the constant vector. Random baselines average 300 pairs of independent random subspaces.

\begin{table}[H]
\centering
\small
\setlength{\tabcolsep}{4pt}
\caption{\textbf{Residual-stream alignment} (mean cosine of principal angles between 11-dimensional subspaces). Not grokked: three seeds each at weight decay 0.1 and 0.3, and the one non-grokked seed at weight decay 1.0.}
\label{tab:subspace_analysis}
\begin{tabularx}{\linewidth}{Xcccc}
\toprule
 & Grokked & \multicolumn{3}{c}{Not grokked} \\
\cmidrule(lr){2-2}\cmidrule(lr){3-5}
Subspace pair & wd 1.0 & wd 0.1 & wd 0.3 & wd 1.0 \\
 & ($n=14$) & ($n=3$) & ($n=3$) & ($n=1$) \\
\midrule
Attention write ($W_O$) vs.\ MLP read ($W_\text{in}$) & $0.923\pm0.035$ & $0.839\pm0.021$ & $0.918\pm0.048$ & $0.947$ \\
MLP read ($W_\text{in}$) vs.\ MLP write ($W_\text{out}$) & $0.330\pm0.050$ & $0.269\pm0.024$ & $0.271\pm0.043$ & $0.218$ \\
MLP write ($W_\text{out}$) vs.\ unembedding read ($W_U$) & $0.885\pm0.078$ & -- & -- & -- \\
MLP read vs.\ key frequencies of $W_E$ & $0.252\pm0.018$ & -- & -- & -- \\
MLP write vs.\ key frequencies of $W_E$ & $0.254\pm0.019$ & -- & -- & -- \\
\midrule
Random baseline & \multicolumn{4}{c}{$0.248\pm0.017$} \\
\bottomrule
\end{tabularx}
\end{table}

\begin{figure}[H]
  \centering
  \begin{subfigure}[b]{0.75\linewidth}
    \centering
    \includegraphics[width=0.8\linewidth]{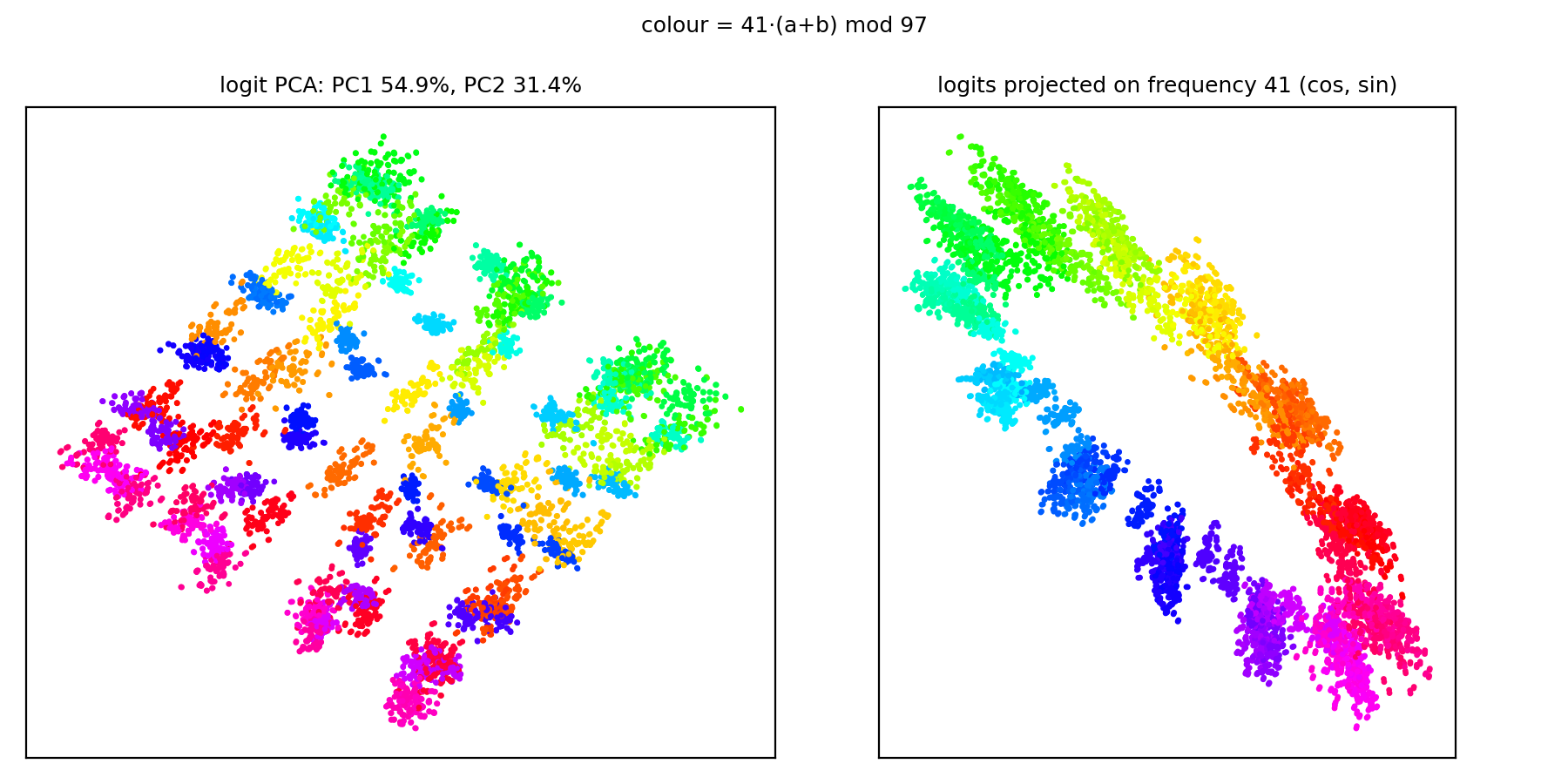}
    \caption{Transformer logits: first two principal components (left) and projection on the dominant logit frequency (right), colored by $41(a+b)\bmod97$.}
  \end{subfigure}
  \begin{subfigure}[b]{0.75\linewidth}
    \centering
    \includegraphics[width=0.8\linewidth]{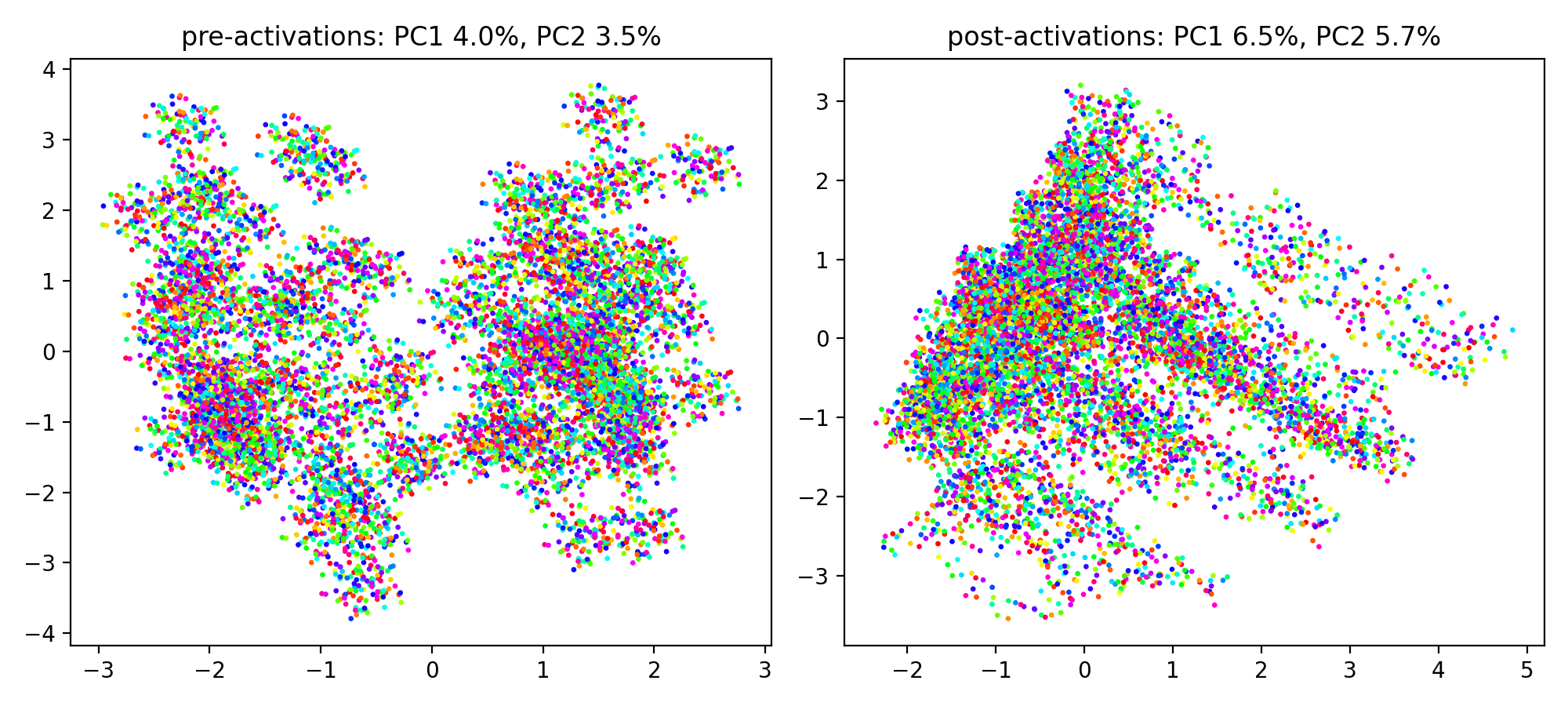}
    \caption{MLP pre-activations (left) and post-activations (right): first two principal components, colored by $(a+b)\bmod97$.}
  \end{subfigure}
  \caption{\textbf{Representational geometry} (seed 0).}
  \label{fig:transformer_causal}
\end{figure}

\subsection{Representational Geometry}

Transformer logits are dominated by a few frequencies, while the MLP's representations are not (Figure~\ref{fig:transformer_causal}). Five frequencies carry 87.9\% of the logit power (seed 0, after removing each input's mean logit). The first two principal components of the transformer's logits carry 54.9\% and 31.4\% of their variance and show clusters of nearby classes, and projected on the strongest frequency ($f^\star=41$) the logits lie on a ring ordered by $41(a+b)\bmod 97$. The first two principal components of the MLP's pre- and post-activations carry only 4.0\%/3.5\% and 6.5\%/5.7\% of the variance and show no ring, consistent with the dozens of frequencies the MLP uses (Section~\ref{sec:comparison}).

\end{document}